%% file: main.tex
\documentclass{article}

\usepackage{iclr2027_conference,times}
\input{math_commands.tex}

\usepackage{amsmath}
\usepackage{amssymb}
\usepackage{booktabs}
\usepackage{float}
\usepackage{graphicx}
\usepackage[caption=false,font=small]{subfig}
\usepackage{microtype}
\usepackage{multirow}
\usepackage[section]{placeins}
\usepackage{tabularx}
\usepackage{longtable}
\usepackage{xcolor}
\usepackage[most]{tcolorbox}
\usepackage{fvextra}
\usepackage{hyperref}
\usepackage{url}
\usepackage{wrapfig}
\usepackage{needspace}
\hypersetup{hidelinks}

\input{prompt_layout.tex}

\newcommand{\todo}[1]{\textcolor{red}{[TODO]}}

\title{Understanding Errors in LLM-Based\\Question Answering over Imperfect Tables}

\input{authors}

\iclrfinalcopy
\hypersetup{
  pdftitle={Understanding Errors in LLM-Based Question Answering over Imperfect Tables},
  pdfauthor={Baowen Zhang, Wei Fan, Ruman Wang, Hangting Ye}
}

\begin{document}
\raggedbottom

\maketitle
\lhead{}
\renewcommand{\headrulewidth}{0pt}

\begin{abstract}
\input{sections/00_abstract}
\end{abstract}

\input{sections/01_introduction}
\Needspace{10\baselineskip}
\input{sections/02_problem_definition}
\input{sections/03_geometry}
\input{sections/04_discovery_not_intervention}
\input{sections/05_end_to_end_validation}
\input{sections/08_related_work}
\input{appendices/appendix_limitations}
\input{sections/09_limitations_conclusion}

\clearpage
\input{backmatter/ai_use_statement}
\input{backmatter/ethics_statement}
\input{backmatter/reproducibility_statement}

\bibliography{references}
\bibliographystyle{iclr2027_conference}

\clearpage
\appendix
\input{appendices/appendix_overview}
\input{appendices/appendix_a_target_coordinates}

\input{appendices/appendix_b_protocols}
\input{appendices/appendix_c_additional_results}
\input{appendices/appendix_f_method_ablations}

\input{appendices/appendix_g_uncertainty}
\input{appendices/appendix_e_prompts}

\end{document}

%% file: math_commands.tex
\usepackage{amsmath,amsfonts,bm}

\def\eqref#1{equation~\ref{#1}}

\def\1{\bm{1}}

\DeclareMathAlphabet{\mathsfit}{\encodingdefault}{\sfdefault}{m}{sl}
\SetMathAlphabet{\mathsfit}{bold}{\encodingdefault}{\sfdefault}{bx}{n}

%% file: prompt_layout.tex
\definecolor{promptink}{HTML}{253541}
\definecolor{promptline}{HTML}{A9B3BA}
\definecolor{promptaccent}{HTML}{397681}
\definecolor{promptsoft}{HTML}{F3F6F7}
\newcommand{\promptusericon}{%
  \tikz[baseline=-0.5ex,x=1ex,y=1ex,line width=.35pt]{%
    \filldraw[draw=promptaccent,fill=promptaccent!8] (0,0) circle (.84);
    \fill[promptaccent] (0,.25) circle (.23);
    \fill[promptaccent] (-.46,-.43) .. controls (-.44,.12) and (.44,.12) .. (.46,-.43) -- cycle;}}
\newcommand{\promptsystemicon}{%
  \tikz[baseline=-0.5ex,x=1ex,y=1ex,line width=.45pt]{%
    \draw[promptink,rounded corners=.4pt] (-.74,-.59) rectangle (.74,.65);
    \draw[promptink] (-.38,.29) -- (-.13,.05) -- (-.38,-.19);
    \draw[promptink] (.02,-.22) -- (.39,-.22);}}
\newcommand{\promptconfigicon}{%
  \tikz[baseline=-0.5ex,x=1ex,y=1ex,line width=.4pt]{%
    \draw[promptink] (-.52,-.64) rectangle (.52,.68);
    \draw[promptink] (-.29,.32)--(.30,.32) (-.29,.02)--(.30,.02) (-.29,-.28)--(.17,-.28);}}
\expandafter\def\csname prompticon@user\endcsname{\promptusericon}
\expandafter\def\csname prompticon@system\endcsname{\promptsystemicon}
\expandafter\def\csname prompticon@config\endcsname{\promptconfigicon}
\expandafter\def\csname promptrole@user\endcsname{User}
\expandafter\def\csname promptrole@system\endcsname{System}
\expandafter\def\csname promptrole@config\endcsname{Configuration}
\newcommand{\promptcardtitle}[2]{%
  \csname prompticon@#1\endcsname\enspace
  \csname promptrole@#1\endcsname\quad{\color{promptline}\textbar}\quad #2}
\tcbset{paperprompt/.style={
  enhanced jigsaw,breakable,
  colback=white,colframe=promptline,colbacktitle=white,coltitle=promptink,
  fonttitle=\normalfont\normalsize\bfseries,
  fontupper=\normalfont\normalsize,
  boxrule=.4pt,titlerule=.35pt,arc=0pt,outer arc=0pt,
  left=8pt,right=8pt,top=5pt,bottom=5pt,
  lefttitle=8pt,righttitle=8pt,toptitle=5pt,bottomtitle=4pt,
  before skip=6pt,after skip=7pt,
  pad at break*=4pt,lines before break=4,
  before upper={\setlength{\parindent}{0pt}\setlength{\parskip}{3pt}\raggedright}
}}
\expandafter\def\csname promptkeep@generic_guidance\endcsname{}
\expandafter\def\csname promptkeep@intervention_preturn\endcsname{}
\expandafter\def\csname promptkeep@qa_user\endcsname{}
\expandafter\def\csname promptkeep@qa_continuation\endcsname{}
\expandafter\def\csname promptkeep@ledger_three_models\endcsname{}

\newcommand{\promptheading}[1]{%
  \par\addvspace{5pt}\noindent{\normalfont\bfseries #1}\par\nobreak\vspace{1pt}}
\newcommand{\promptinline}[1]{{\fontencoding{T1}\fontfamily{ptm}\selectfont #1}}
\newcommand{\promptitem}[3]{%
  \par\begingroup\setlength{\parskip}{1.4pt}\setlength{\leftskip}{#1em}%
  \hangindent=1.0em\hangafter=1\noindent
  \makebox[1em][l]{#2}#3\par\endgroup}
\newcommand{\promptfield}[2]{%
  \par\noindent\textbf{#1}\par\nobreak
  {\color{promptaccent}#2}\par}
\newcommand{\promptcodefile}[1]{%
  \par\vspace{1pt}\noindent\begin{minipage}{\linewidth}%
  \fontencoding{T1}\fontfamily{ptm}\selectfont
  \VerbatimInput[fontsize=\small,fontfamily=ptm,baselinestretch=.95,
    breaklines=true,breakanywhere=false,breaksymbolleft={},
    breakindent=1em,xleftmargin=8pt,
    frame=leftline,framerule=.6pt,rulecolor=\color{promptline},framesep=6pt]{#1}%
  \end{minipage}\par\vspace{1pt}}

%% file: authors.tex
\author{%
Baowen Zhang$^{1}$, Wei Fan$^{2}$, Ruman Wang$^{3}$,
Hangting Ye$^{4}$\thanks{Corresponding author.}\\
University of Wisconsin--Madison$^{1}$\\
University of Auckland, Auckland, New Zealand$^{2}$\\
Liaoning Provincial People's Hospital$^{3}$\\
School of Artificial Intelligence, Jilin University$^{4}$\\
\texttt{bzhang599@wisc.edu}, \texttt{wei.fan@auckland.ac.nz},\\
\texttt{ruman.wang@outlook.com}, \texttt{yeht22@mails.jlu.edu.cn}
}

%% file: sections/00_abstract.tex
Answering questions over imperfect tables requires handling errors that can affect the answer. We investigate two challenges for large language models (LLMs): whether error discovery depends on where errors appear in a table, and whether providing their locations is sufficient for accurate question answering (QA). Using human-reviewed instances from RADAR-T, we conduct controlled studies across three LLMs by varying row order and comparing original, error-marked, and repaired tables. First, reordering rows changes error discovery even when the table contents and gold answer remain unchanged. During direct inspection, LLMs are more likely to discover all rows containing relevant errors when these rows appear later in the table or are grouped more closely together. Second, providing verified error locations alone is insufficient for accurate QA: with code execution, accuracy on repaired tables exceeds that on error-marked tables by 39.0--59.1 percentage points across the three LLMs. As a practical application of these findings, we combine error discovery across shuffled table views with explicit guidance for verifying and handling the reported errors in a simple workflow, Geometry-Balanced Discovery and Intervention (GBDI). On RADAR-T, GBDI improves QA accuracy by 3.8--18.5 percentage points over a code-agent baseline across five LLMs (paired 95\% confidence intervals exclude zero for four), at the cost of additional inference. These results highlight the importance of both reliable error discovery and effective error handling in QA over imperfect tables.
Code is available at
\mbox{\urlstyle{same}\url{https://github.com/645-t/GBDI-ICLR-2027}}.

%% file: sections/01_introduction.tex
\section{Introduction}
\label{sec:introduction}

Tabular data organize records into rows and features into columns, with each cell representing a feature value whose interpretation depends on the column semantics and data type~\citep{wickham2014tidy,hulsebos2019sherlock}. This structured representation supports data analysis across domains such as healthcare~\citep{johnson2023mimiciv}, finance~\citep{chen2021finqa}, and materials design~\citep{choudhary2020jarvis}. Table question answering (TQA)~\citep{pasupat2015compositional,herzig2020tapas} enables users to access and analyze tabular data through natural-language questions, which often involve operations such as filtering, comparison, and aggregation~\citep{wang2024chain,zhu2021tatqa}.
Recent methods use large language models (LLMs) to interpret questions and reason over tables, often with the help of executable code or explicit table operations~\citep{jiang2023structgpt,wang2024chain,gao2023pal}. Progress in reasoning over a supplied table, however, does not by itself establish reliability when its contents are imperfect. Real-world tables can contain errors such as missing values and inconsistent formats~\citep{gu2025radar}. Understanding how these errors affect LLM-based analysis is therefore essential for reliable TQA beyond clean inputs.

Answering questions over imperfect tables involves two connected demands: discovering errors that matter to the question and deciding how to handle them before computing an answer. LLMs can critique and revise outputs through iterative feedback~\citep{madaan2023selfrefine}, but reasoning corrections remain unreliable without external feedback~\citep{huang2024selfcorrect}. Neither result establishes reliable error handling in tables. In prose, word order contributes directly to meaning; in a table, a value must be interpreted through its column, its relations to other values, and the requested computation. For many questions, the displayed order of records is incidental. RADAR shows that data imperfections substantially degrade LLM-based analysis~\citep{gu2025radar}, motivating a closer question: where do these systems struggle---in discovering relevant errors, or in translating their discovery into correct answers? We investigate two corresponding challenges.

\noindent\textbf{Challenge 1: Discovering Errors.}
LLMs receive textual table representations as token sequences, raising the possibility that error discovery depends on presentation order even when that order is irrelevant to the answer. Prior work shows that semantically equivalent table arrangements can change final predictions~\citep{liu2024rethinking,dong2026power}. We ask whether this sensitivity also affects which errors are discovered before answering. Specifically, can the same errors be overlooked when their rows appear earlier in the table or farther apart? We call rows containing errors that can affect the answer \emph{critical rows}. Reliable systems should discover these rows regardless of their placement whenever reordering preserves the meaning of the question and gold answer.

\noindent\textbf{Challenge 2: Handling Errors.}
Discovering an erroneous value does not determine a valid replacement or how that value should enter the computation. A correction must be supported by the available table evidence; when a required value cannot be recovered, the affected record may need to be excluded under the task rules. An error notification therefore leaves substantive decisions unresolved. We ask whether explicitly marking the relevant errors is enough to recover accurate answers, or whether a substantial gap remains between knowing where the errors are and answering from a table in which they have already been handled.

To investigate these challenges, we conduct two controlled studies on human-reviewed RADAR-T instances across three systems. The first uses answer-preserving changes in critical-row placement, holding all row contents, the question, and the gold answer fixed. On 60 instances, we vary where critical rows appear; on the 59 instances with multiple critical rows, we also vary their spacing at controlled mean positions. Complete discovery is higher when critical rows are placed near the back rather than the front. Averaged over the tested mean positions, Wide layouts yield lower complete discovery rates than Compact layouts. Thus, answer-preserving changes in presentation can change whether relevant errors are discovered.

The second study compares answers on original tables, tables with verified errors marked, and tables with reviewed repairs already applied, using 154 reviewed instances. Error-marked inputs yield smaller and uncertain changes in accuracy, whereas providing repaired tables yields much larger gains for code-assisted QA. Across the three systems, providing repaired rather than marked tables raises accuracy by 39.0--59.1 percentage points. Together, the studies reveal both a sensitivity in error discovery and a substantial difficulty that remains after error locations are supplied.

To illustrate the practical implications of these findings, we combine inspection across shuffled table views with explicit guidance for verifying and handling reported errors in a simple workflow, Geometry-Balanced Discovery and Intervention (GBDI). The workflow aggregates candidate errors from multiple views, checks them against the visible table, and applies supported actions before final QA. On all 313 RADAR-T instances, GBDI raises observed accuracy over the Code Agent baseline by 3.8--18.5 percentage points across five systems; paired 95\% confidence intervals exclude zero for four. Separate controls compare shuffled and repeated inspections at the same call budget and vary intervention guidance while holding the discovery evidence fixed.

Our contributions are:
\begin{itemize}
    \item We show that the discovery of question-relevant errors depends on their placement, even when row permutations preserve the table contents and gold answer.
    \item We distinguish marked inputs from inputs with completed repairs, showing that providing error locations without repairs leaves a substantial gap in final QA accuracy.
    \item We demonstrate a practical application of these findings through GBDI, a simple workflow that combines varied table views with explicit error-handling guidance to improve final QA.
\end{itemize}

%% file: sections/02_problem_definition.tex
\section{Problem Formulation}
\label{sec:problem}

Following RADAR~\citep{gu2025radar}, a base task is a pair $(T_s,Q)$ of a clean
source table and a natural-language question. RADAR introduces one of five artifact
types---missing values, bad values, outliers, formatting errors, and logical
inconsistencies---into a table $T$ derived from $T_s$; clean variants omit this
perturbation. Each TQA instance is then a triple $(T,Q,A)$, where $T$ is the
resulting observed table and $A$ is the gold answer computed from the
corresponding recovered table. The system receives $T$ and $Q$ and predicts
$\hat A$, which we evaluate using the RADAR answer-scoring rules.

\subsection{Critical Cells}

For instance $i$, let $T_i$ denote its observed table and $c$ a cell in it.
Let $E_i$ denote the set of cells in $T_i$ that contain a human-verifiable
error, and let $D_i(Q_i)$ denote the set of cells whose values can affect the
answer to $Q_i$, for example through filtering, comparison, or aggregation.
The set of critical cells is their intersection:
\begin{equation}
    \mathcal{C}_i
    =
    E_i \cap D_i(Q_i).
    \label{eq:critical-cell}
\end{equation}
A critical cell is both erroneous and relevant to the current question.

A table may contain errors that do not affect the answer to $Q_i$. These cells
belong to $E_i$ but not to
$\mathcal{C}_i$. Conversely, a cell may be relevant to the
question but contain no error; such a cell belongs to $D_i(Q_i)$ but not to
$\mathcal{C}_i$. Our error-discovery analysis focuses on these
critical cells and the rows that contain them.

Because our row-order experiments operate on rows, we also define a critical
row as any row containing at least one critical cell:
\begin{equation}
    \mathcal{R}_i
    =
    \left\{
        r\in T_i
        \;:\;
        \exists\, c\in r
        \text{ with }
        c\in\mathcal{C}_i
    \right\}.
    \label{eq:critical-row}
\end{equation}
Thus, $\mathcal{C}_i$ identifies the error locations that matter
for the question, while $\mathcal{R}_i$ identifies the rows that
contain them.

\subsection{Intervention}

Discovering a critical row does not by itself determine how to handle its
errors. For each critical row $r\in\mathcal{R}_i$, let $a_i^{\mathrm H}(r)$
denote its human-reviewed intervention:
\begin{equation}
    a_i^{\mathrm H}(r)=
    \begin{cases}
        \text{\textsc{Edit}}, &
        \text{if every required correction can be reliably recovered},\\
        \text{\textsc{Drop}}, &
        \text{otherwise}.
    \end{cases}
    \label{eq:human-action}
\end{equation}

\textsc{Edit} replaces the implicated erroneous values with supported
corrections. \textsc{Drop} excludes the affected row from the computation when
at least one required correction cannot be reliably recovered.

\paragraph{Human-reviewed reference.}
To evaluate discovery and intervention against a common reference, we review
154 RADAR-T instances. Reviewers inspect each table and question to identify
critical cells and their containing rows, and verify the supported
\textsc{Edit} replacements or \textsc{Drop} decisions. These annotations
provide reference rows for discovery evaluation in Section~\ref{sec:geometry}
and the error markings and completed repairs compared in
Section~\ref{sec:intervention-motivation}. Appendix~\ref{app:target-coordinates}
describes the review procedure, coverage, and subset selection.

\subsection{Answer-Preserving Row Permutations}

To study the effect of row order, let $\pi$ denote a permutation of the data
rows of $T_i$ with the header fixed. We consider questions that do not refer
directly to row positions, such as the first or last row. Values with semantic
meaning, such as dates or ranks, remain attached to their original rows.

Let $\operatorname{Answer}(Q,T)$ denote the gold answer for question $Q$ and
observed table $T$, computed from the corresponding recovered table. Under these
conditions, changing the displayed row order preserves this gold answer:
\begin{equation}
    \operatorname{Answer}(Q_i,\pi(T_i))
    =
    \operatorname{Answer}(Q_i,T_i)
    =
    A_i.
    \label{eq:row-order-invariance}
\end{equation}

The critical cells also remain unchanged; only their displayed row positions
change. We can therefore vary where critical rows appear while holding the
underlying data, question, and gold answer fixed.

\subsection{Complete Discovery Rate}
\label{sec:complete-discovery-rate}

For a fixed system--mode pair, let $\widehat{\mathcal R}_{i,\pi}$ denote the
nominated row-ID set from view $\pi$, with valid IDs mapped to canonical row
identities from the original table. For a collection of evaluated views
$\mathcal V$ with nonempty critical-row sets, the complete discovery rate (CDR) is
\begin{equation}
 \operatorname{CDR}(\mathcal V)
 =\frac{100}{|\mathcal V|}
 \sum_{(i,\pi)\in\mathcal V}
 \mathbf{1}\!\left[\mathcal{R}_i
 \subseteq\widehat{\mathcal R}_{i,\pi}\right].
 \label{eq:complete-discovery-rate}
\end{equation}
A view counts as successful only if the system discovers every critical row
in the reference set. Missing any critical row counts as a
discovery failure. CDR measures complete coverage, so additional nominated rows
do not reduce the score. Invalid submissions are treated as empty predictions.
Appendix~\ref{app:discovery-metrics} complements CDR with row-level recall,
precision, and F1, the number of nominated rows, and the false-positive rate on
instances with no critical rows.

In Section~\ref{sec:geometry}, we use these controlled row permutations to
study whether the position and dispersion of critical rows affect error
discovery. In Section~\ref{sec:intervention-motivation}, we instead compare
answers obtained from the original table, a table with reviewed errors marked,
and a table with reviewed repairs applied.

%% file: sections/03_geometry.tex
\section{Discovering Errors: Geometric Sensitivity}
\label{sec:geometry}

Imperfect TQA requires a system to discover the critical rows that contain
errors capable of changing the answer. For questions that do not depend on
displayed row order, a row permutation changes neither the critical-row set nor
the gold answer. We test whether systems respect this invariance by moving
the same human-reviewed critical rows through answer-preserving permutations.
We vary two properties of their placement: position and, for instances with
multiple critical rows, dispersion.

\subsection{Setup: Position and Dispersion}
\label{sec:target-set-geometry}
\label{sec:controlled-geometry}

We describe critical-row placement by its average position in the table and
by the spacing between critical rows. For instance $i$, let $n_i\geq2$ be the
number of data rows and $k_i=|\mathcal R_i|\geq1$ the number of critical rows.
We index the critical rows by $j\in\{1,\ldots,k_i\}$, independently of their
displayed order. In a row-permuted view $\pi$, let $x^\pi_{ij}$ be the zero-based
displayed row index of critical row $j$ divided by $n_i-1$. The first and last
rows thus have normalized positions 0 and 1, respectively.

\begin{wrapfigure}[16]{r}{0.27\textwidth}
\vspace{-0.7em}
\centering
\includegraphics[width=\linewidth]{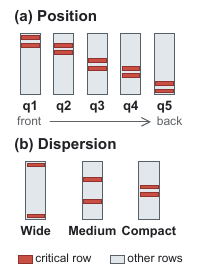}
\par\vspace{-0.6em}
\setlength{\abovecaptionskip}{4pt}
\caption{Position and dispersion.}
\label{fig:geometry-schematic}
\vspace{-0.8em}
\end{wrapfigure}

\noindent\textbf{Position.} We measure the average location of the critical rows
by their mean normalized position:
\begin{equation}
 \mu_{i,\pi}=\frac{1}{k_i}\sum_{j=1}^{k_i}x^\pi_{ij}.
 \label{eq:mean-position}
\end{equation}
Smaller values indicate that the critical rows lie nearer the beginning of the
table on average; larger values indicate that they lie nearer the end.

\noindent\textbf{Dispersion.} For $k_i\geq2$, we average the distance from each
critical row to its nearest other critical row:
\begin{equation}
 \sigma_{i,\pi}=\frac{1}{k_i}\sum_{j=1}^{k_i}
 \min_{\substack{1\leq j'\leq k_i\\j'\neq j}}
 \left|x^\pi_{ij}-x^\pi_{ij'}\right|.
 \label{eq:target-dispersion}
\end{equation}
Larger values indicate greater average distance to the nearest other critical
row. Figure~\ref{fig:geometry-schematic} illustrates both quantities. The
position experiment varies $\mu_{i,\pi}$, whereas the dispersion experiment
varies $\sigma_{i,\pi}$ while holding $\mu_{i,\pi}$ fixed within each comparison.

\WFclear
\input{figures/geometry_figure_pair}

\subsection{Experimental Design}

We use 60 instances from the human-reviewed reference, with 12 from each
of the five error types, covering 29 base tasks and 21 source tables.
Instances are selected using annotation metadata and layout feasibility
(Appendix~\ref{app:geometry-cohort-selection}). Each comparison uses the
same question, prompt, cell values, and critical rows.

\paragraph{Position experiment.}
All 60 instances are used. For each instance, we keep the critical rows
together as a contiguous block and vary their mean position across five
bands, q1--q5, from the front to the back of the table
(Figure~\ref{fig:geometry-schematic}(a)). The remaining rows retain their
relative order as the block moves. We repeat these five placements under
three fixed orders of the remaining rows, yielding
$60\times3\times5=900$ views per system--mode pair.

\paragraph{Dispersion experiment.}
The dispersion experiment covers the 59 instances with multiple critical
rows. The remaining instance has a single critical row, for which spacing
between critical rows cannot be measured. For each instance, we construct three fixed orders of the
remaining rows and test three target mean positions: 25\%, 50\%, and 75\%.
At each target position and fixed row order, we compare Compact, Medium,
and Wide layouts with increasing spacing between critical rows
(Figure~\ref{fig:geometry-schematic}(b)). The individual critical-row positions vary
across these layouts, while their mean position remains fixed within each
comparison. This gives
$59\times3\times3\times3=1{,}593$ views per system--mode pair.
The mean realized $\sigma$ is 0.008, 0.068, and 0.127 for Compact, Medium,
and Wide, respectively. Appendix~\ref{app:geometry-protocol} describes
the placement procedure.

\paragraph{Systems and inference.}
We evaluate Qwen Plus, MiniMax-M2.5, and GPT-5 Mini using the Direct and
Code Agent modes adapted from RADAR~\citep{gu2025radar}. In this experiment,
both modes identify rows that may contain errors relevant to the question.
Direct reports these rows in one response. The Code Agent receives the
same complete table and question and may inspect the table with read-only
Python, using up to five responses including its final submission.

Figure~\ref{fig:geometry-main} reports CDR for both experiments and inference
modes. Numerical results and uncertainty estimates appear in
Appendices~\ref{app:geometry-marginals} and~\ref{app:uncertainty}.
We also evaluate position and dispersion effects on missingness instances
constructed from WikiSQL following the RADAR missingness setting
(Appendix~\ref{app:external-wikisql}).

\subsection{Results}

CDR is higher when critical rows appear near the end of the table (q5) than
near the beginning (q1). With Direct, Qwen rises from 61.1\% to 72.8\%,
MiniMax from 47.8\% to 59.4\%, and GPT from 43.9\% to 60.6\%
(Figure~\ref{fig:geometry-main}(a)). The Code Agent
shows the same pattern, with smaller increases for all three systems
(Figure~\ref{fig:geometry-main}(b)).

Wide layouts yield lower CDR than Compact layouts for all three systems
using Direct (Figure~\ref{fig:geometry-main}(a)).
Averaged over the three tested values of $\mu$, the decrease is 16.2 points for
Qwen, 9.6 for MiniMax, and 7.9 for GPT. The same direction holds at each tested
$\mu$ for every system under Direct
(Appendix~\ref{app:joint-geometry}).

These controlled comparisons identify presentation sensitivity in discovering
question-relevant errors while holding the question, table contents, and gold
answer fixed.

\FloatBarrier

%% file: figures/geometry_figure_pair.tex
\begin{figure}[!t]
\centering
\captionsetup[subfloat]{position=bottom,justification=centering,singlelinecheck=false,
margin={16bp,0pt},font={small,bf},captionskip=3pt,nearskip=0pt,farskip=0pt}
\subfloat[Direct\label{fig:geometry-direct}]{%
\includegraphics[width=0.49\linewidth]{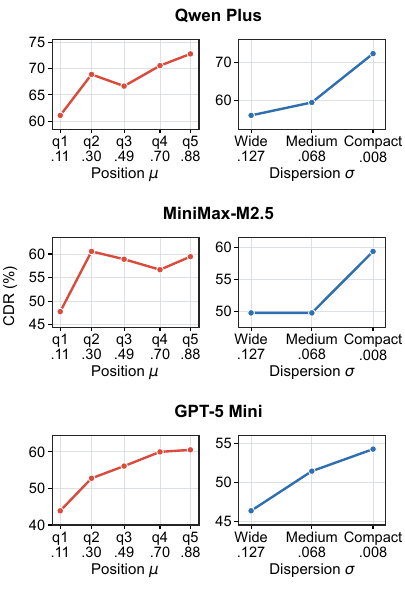}}\hfill
\subfloat[Code Agent\label{fig:geometry-code-agent}]{%
\includegraphics[width=0.49\linewidth]{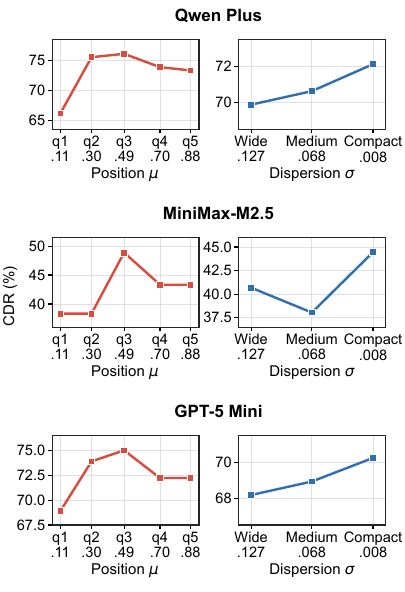}}
\caption{Complete discovery rate under (a) Direct and (b) Code Agent.
In each subfigure, rows correspond to systems; the left column varies position
(q1--q5, front to back), and the right column varies dispersion (Wide to Compact). Tick labels report
mean realized $\mu$ or $\sigma$.}
\label{fig:geometry-main}
\end{figure}

%% file: sections/04_discovery_not_intervention.tex
\section{Handling Errors: Discovery Alone Is Insufficient}
\label{sec:intervention-motivation}
\label{sec:capability}

Even when relevant errors are discovered, their locations do not specify how
they should be handled. The model must still determine supported corrections
or exclusions. We therefore compare error marking with completed,
human-reviewed repairs to examine whether supplying error locations is
sufficient for accurate QA.

\subsection{Experimental Setup}

We evaluate Qwen Plus, MiniMax-M2.5, and GPT-5 Mini in both Direct and
Code Agent modes on all 154 instances in the human-reviewed reference.
Under each table condition, the model answers the question, and we evaluate
final-answer accuracy.

We compare three table states, with their Direct prompts shown in
Figure~\ref{fig:table-state-prompts}.
Raw uses the original table. Discovery (Disc, error-marked) wraps each reviewed critical
cell in \texttt{<<ERROR:value>>}, retaining the original value inside the marker.
Action (Act, repaired) applies the reviewed replacements and removes the reviewed
\textsc{Drop} rows.

\input{figures/table_state_prompts}

Discovery asks the model to handle the marked errors, whereas Action provides
a repaired table for answering. All conditions use the same question;
none supplies the gold answer.

For instance $i$, let $\mathcal P_i^{\mathrm H}$ denote its repair plan:
supported replacements for erroneous cell values in \textsc{Edit} rows and
the \textsc{Drop} rows to remove. With $\operatorname{Mark}$ and
$\operatorname{Repair}$ denoting the marking and repair operations, respectively,
the three table states are:
\begin{equation}
 \begin{aligned}
 T_i^{\mathrm{Raw}} &= T_i, \\
 T_i^{\mathrm{Disc}} &= \operatorname{Mark}(T_i,\mathcal C_i), \\
 T_i^{\mathrm{Act}} &= \operatorname{Repair}(T_i,\mathcal P_i^{\mathrm H}).
 \end{aligned}
 \label{eq:table-state-construction}
\end{equation}
\Needspace{6\baselineskip}
\subsection{Results}

Providing repaired rather than error-marked tables raises observed QA accuracy
across all three systems and both inference modes
(Figure~\ref{fig:intervention-table-states}). The increase is particularly
pronounced under the Code Agent: 39.0--59.1 percentage points, with all paired
intervals excluding zero. By comparison,
changes from error marking relative to Raw inputs are smaller and statistically
uncertain. Detailed results and paired intervals are reported in
Appendix~\ref{app:table-state-contrasts},
Table~\ref{tab:intervention-information-uncertainty}.

\input{figures/intervention_table_states}

These results distinguish discovering relevant errors from resolving them for
question answering. Providing reviewed repairs yields higher QA accuracy than
providing verified error locations alone, motivating an explicit intervention
stage after discovery.

%% file: figures/table_state_prompts.tex
\begingroup
\setlength{\intextsep}{6pt}
\begin{figure}[H]
\centering
\includegraphics[width=\linewidth]{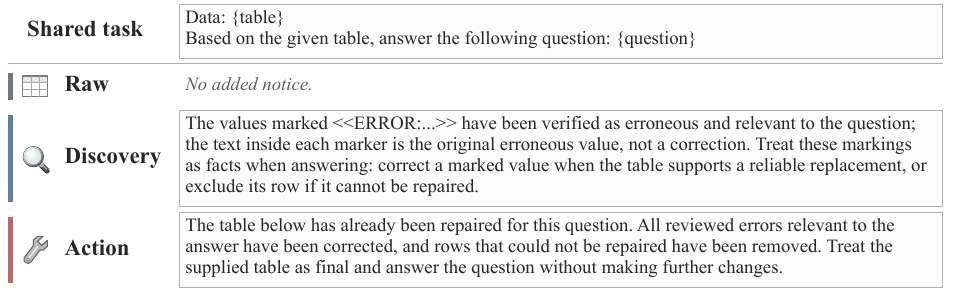}
\caption{The shared Direct task template (top) and three input conditions (below).
Each notice, when present, precedes the task in the actual prompt.
\texttt{\{table\}} receives the corresponding table; \texttt{\{question\}} is unchanged.
Code Agent uses the same notices with its own task template. Full prompts appear
in Appendix~\ref{app:prompts}.}
\label{fig:table-state-prompts}
\end{figure}
\endgroup

%% file: figures/intervention_table_states.tex
\begin{figure}[!t]
\centering
\includegraphics[width=\linewidth]{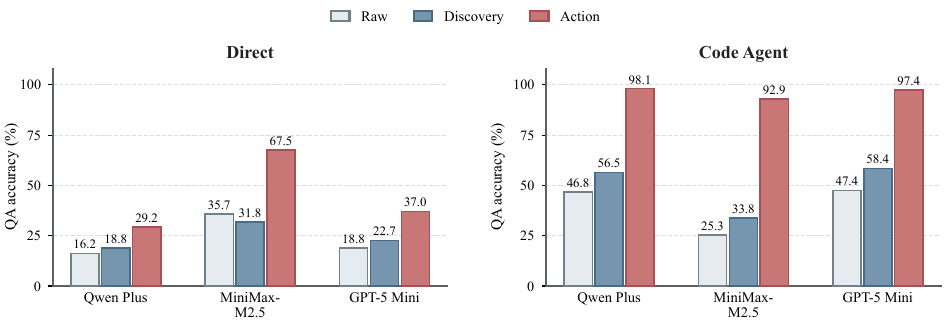}
\caption{Final-answer accuracy (\%) on 154 reviewed RADAR-T instances.
\textsc{Raw} uses the original table, \textsc{Discovery} (error-marked) marks
critical cells, and \textsc{Action} (repaired) provides the table with reviewed
cell corrections and \textsc{Drop} row removals already applied. Paired intervals are reported in
Appendix~\ref{app:uncertainty}.}
\label{fig:intervention-table-states}
\end{figure}

%% file: sections/05_end_to_end_validation.tex
\section{Geometry-Balanced Discovery and Intervention}
\label{sec:end-to-end-validation}
\label{sec:experiments}

GBDI puts the two findings into practice by combining discovery across
shuffled table views with explicit guidance for verifying and handling
candidate errors.
We evaluate how view randomization affects discovery, how intervention
guidance affects QA with discovery reports held fixed, and how the complete
workflow performs on final QA.

\subsection{GBDI Workflow}
\label{sec:gbdi-workflow}

\paragraph{Multi-view discovery.}
As shown in Figure~\ref{fig:gbdi-workflow}, GBDI constructs $M$ views by
independently sampling row permutations uniformly, keeping columns and cell
values unchanged. Each view is inspected in a separate Direct context with
the same question, prompt, and decoding configuration. Reported candidate
errors are mapped back to the original rows and merged into a single report,
termed the union ledger. We call this discovery procedure Random-$M$, where
$M$ is a hyperparameter set to 5 by default.

\paragraph{Guided intervention.}
The Code Agent receives the original table, question, union ledger, and
Intervention Skill: prompt instructions for verifying candidates against the
question and visible table evidence before choosing actions. The Skill
provides guidance on value recovery, formatting, and execution checks. It
instructs the agent to use \textsc{Edit} when the evidence supports a unique
correction, \textsc{Drop} when a question-relevant row is verified to be unusable
and no supported correction is available, and \textsc{No-op} when an
intervention is not justified.

The agent applies the selected actions to a working copy and then answers the
original question in the same conversation and Python state. Appendices
\ref{app:protocols} and~\ref{app:prompts} provide the prompts and implementation
details.

\input{figures/gbdi_workflow}

\Needspace{22\baselineskip}
\subsection{GBDI Evaluation}
\label{sec:setup}
\label{sec:ablation-study}

\input{figures/main_results_gbea}

We evaluate Qwen Plus, GLM-4.5-Air, MiniMax-M2.5, GPT-5 Mini, and DeepSeek-V4.1-Flash on all 313 RADAR-T instances from 53 base tasks and 27 source tables.
The Direct and Code Agent baselines use the released row order. With the
default $M=5$, GBDI adds five inspection calls and reserves one response for
intervention. We compare Random-5 with five independent inspections of the
original row order (Repeated-5); paired discovery and QA results are reported
in Appendices~\ref{sec:core-ablations} and~\ref{app:gbdi-discovery-qa}. For QA,
paired 95\% confidence intervals use source-table cluster
resampling~\citep{field2007bootstrap}; prompts and execution details appear
in Appendix~\ref{app:prompts}.

\paragraph{QA accuracy.}
Relative to Code Agent, GBDI raises QA accuracy point estimates by
3.8--18.5 percentage points across all five systems
(Figure~\ref{fig:main-gbea-results}).

\paragraph{Intervention guidance.}
With the Qwen Plus Random-5 discovery reports held fixed, the Intervention
Skill improves QA by 6.4 percentage points over general guidance on allowed
operations.
Removing formatting rules reduces accuracy;
the effects of removing recovery rules or execution checks are less certain
(Appendices~\ref{sec:core-ablations} and~\ref{app:skill-prompt-ablations}).

\paragraph{QA after complete discovery.}
Even when all critical rows are discovered, incorrect answers remain across
Qwen, GLM, and MiniMax. Among instances with complete discovery, replacing
generic guidance with Skill while keeping the discovery reports fixed changes
more answers from incorrect to correct than from correct to incorrect
(Appendix~\ref{app:discovery-answer-decomposition}).

\paragraph{Sensitivity to view count.}
For Qwen Plus, CDR shows diminishing returns as $M$ increases to five, while
final QA at $M=5$ improves over using the same Skill without discovery reports.
Individual QA increments remain uncertain
(Appendices~\ref{app:view-budget-support} and~\ref{app:view-count-qa}).

%% file: figures/gbdi_workflow.tex
\begin{figure}[!htb]
    \centering
    \includegraphics[width=\linewidth]{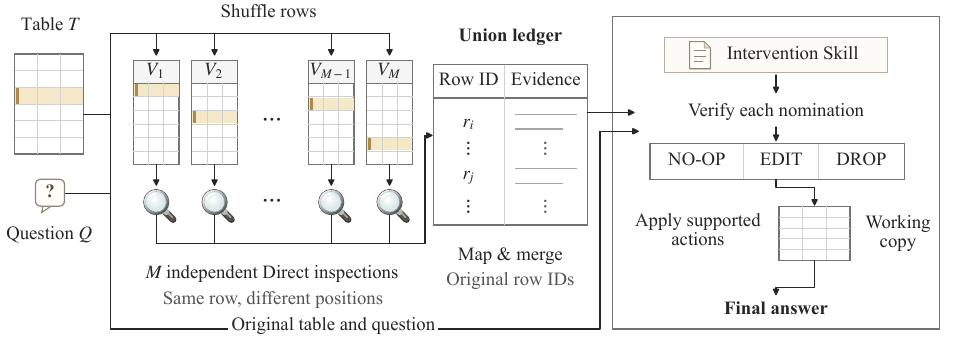}
    \caption{GBDI workflow. $M$ independently shuffled views of the input table are inspected
    in separate Direct contexts. Candidate errors and supporting evidence are mapped
    to the original rows and merged. The Code Agent verifies each nomination, applies
    supported actions to a working table, and computes the final answer. The
    highlighted row illustrates a change in position across views.}
    \label{fig:gbdi-workflow}
\end{figure}

%% file: figures/main_results_gbea.tex
\begin{wrapfigure}[17]{r}{0.46\textwidth}
    \vspace{-0.7em}
    \centering
    \includegraphics[width=\linewidth]{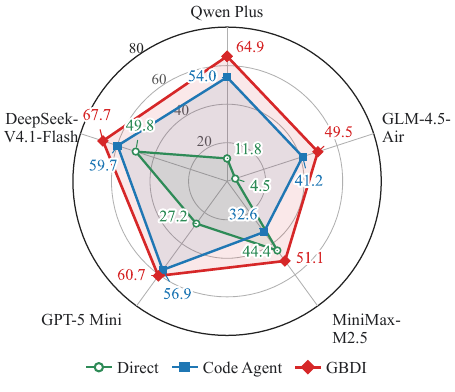}
    \par\vspace{-4pt}
    \caption{QA accuracy (\%) on 313 RADAR-T instances.
    Numerical results and paired CIs are in
    Appendix~\ref{app:gbdi-discovery-qa}, Table~\ref{tab:main-gbea-results}.}
    \label{fig:main-gbea-results}
\end{wrapfigure}

%% file: sections/08_related_work.tex
\Needspace{8\baselineskip}
\section{Related Work}
\label{sec:related-work}

\paragraph{LLMs in TQA.}
StructGPT retrieves structured information through specialized interfaces,
while Chain-of-Table constructs intermediate tables through successive
reasoning operations~\citep{jiang2023structgpt,wang2024chain}. Earlier models
use joint text--table representations, cell selection and aggregation, or
execution-oriented pretraining~\citep{herzig2020tapas,yin2020tabert,liu2022tapex}.
These approaches develop representations or reasoning pipelines for deriving
answers from supplied tables. We study how question-relevant errors are
discovered and handled, using RADAR's Direct and Code Agent baselines for
controlled comparisons within the same systems and inference interfaces.

\paragraph{Robustness for LLMs.}
General LLM robustness studies examine adversarial prompt
perturbations~\citep{zhu2024promptrobust} and distractions from irrelevant
context~\citep{shi2023distracted}. For TQA, RobuT evaluates question and table
perturbations and uses adversarial data augmentation for model
fine-tuning~\citep{zhao2023robut}. FREB-TQA tests retrieval and reasoning
robustness, including the placement of answer-containing rows, and investigates
fine-tuning on transposed tables~\citep{zhou2024freb}. AutoDCWorkflow generates
data-cleaning workflows for a specified analysis purpose; its benchmark
evaluates answer, data, and workflow quality~\citep{li2024autodcworkflow}.
Our controlled studies examine the discovery and handling of erroneous cells
relevant to the question. We build
on RADAR's imperfect-table setting, which requires supported corrections or
record exclusions~\citep{gu2025radar}, and distinguish the effects of marking verified errors from those of supplying completed repairs.

\paragraph{Order Sensitivity in LLMs.}
TableFormer addresses permutation robustness through a pretrained table--text
encoder~\citep{yang2022tableformer}.
Rethinking Tabular Data Understanding with Large Language
Models~\citep{liu2024rethinking} examines table structure variations and
combines textual and symbolic reasoning through answer aggregation. Long-context studies
examine how evidence position and spacing affect retrieval and
reasoning~\citep{liu2024lost,tian2025distance}.
The Power of Order~\citep{dong2026power} evaluates random row and column
permutations and introduces a gradient-based attack that searches for
permutations that degrade final-answer alignment. Its objective is to expose
answer failures through adversarial inputs. Our target is the discovery of
erroneous cells that matter to a question: the model must assess both whether
a cell value is erroneous and whether that error is relevant to the requested
computation. We measure this discovery explicitly and separately examine the
QA gap between supplied error locations and completed repairs.

%% file: appendices/appendix_limitations.tex
\section{Limitations}
\label{sec:limitations}

First, our controlled diagnostics use imperfect tables, where explicit rows and columns
support precise error localization. For the questions studied here, row
permutations preserve the gold answer, enabling controlled tests of error
discovery as row placement varies. Extending these diagnostics to text, images,
or multimodal inputs requires definitions of question-relevant errors,
answer-preserving transformations, and supported interventions,
as well as reliable evaluation annotations.

Second, our experiments access LLMs through black-box interfaces, so connecting
the observed effects to specific internal representations requires further
investigation. Combining the controlled diagnostics with analyses of internal
activations could help explain how models represent error evidence and use it
during discovery, intervention, and answer generation.

%% file: sections/09_limitations_conclusion.tex
\section{Conclusion}

We investigated error discovery and error handling in LLM-based TQA over
imperfect tables. We used answer-preserving changes in row position and dispersion to probe
error discovery under controlled conditions. Complete discovery is higher for
back than front placements and, averaged over the tested mean positions, for
Compact than Wide layouts.
Providing verified error locations alone leaves a substantial QA
accuracy gap relative to tables with reviewed repairs already applied.
As a practical application of these findings, GBDI combines discovery across
shuffled table views with guidance for verifying and handling reported errors.
It achieves higher observed QA accuracy than both Direct and Code Agent
across all five systems. These findings highlight two complementary challenges: discovering
question-relevant errors and handling them using the available evidence.

%% file: backmatter/ai_use_statement.tex
\subsection*{AI use statement}

We used LLMs to check grammar, format LaTeX, and clarify technical descriptions
in this manuscript. The scientific contributions and conclusions are based
on research conducted by the authors.

%% file: backmatter/ethics_statement.tex
\subsection*{Ethics statement}
We confirm that this research complies with all applicable ethical guidelines and
does not present any ethical issues.

%% file: backmatter/reproducibility_statement.tex
\subsection*{Reproducibility statement}

Appendix~\ref{app:target-coordinates} documents the human-reviewed reference
and annotation procedure. Appendix~\ref{app:protocols} specifies the
experimental protocols, model configurations, and scoring rules.
Appendix~\ref{app:uncertainty} describes confidence interval estimation,
and Appendix~\ref{app:prompts} provides the prompts and output schemas
used in our experiments. Code is available at
\url{https://github.com/645-t/GBDI-ICLR-2027}.

%% file: appendices/appendix_overview.tex
\section*{Appendix Contents}

\begingroup
\normalsize
\setlength{\parindent}{0pt}
\setlength{\parskip}{0pt}
\newcommand{\appcontentssection}[1]{%
  \par\addvspace{0.7\baselineskip}%
  \noindent\hyperref[#1]{%
    \bfseries\makebox[2.4em][l]{\ref*{#1}}%
    \nameref*{#1}\hfill\makebox[2em][r]{\pageref*{#1}}}\par
}
\newcommand{\appcontentssubsection}[1]{%
  \noindent\hspace*{1.4em}\hyperref[#1]{%
    \makebox[2.8em][l]{\ref*{#1}}\nameref*{#1}%
    \nobreak\hspace{0.5em}\dotfill\nobreak\hspace{0.5em}%
    \makebox[2em][r]{\pageref*{#1}}}\par
}

\appcontentssection{app:target-coordinates}
\appcontentssubsection{app:reference-coverage}
\appcontentssubsection{app:human-review-protocol}
\appcontentssubsection{app:geometry-cohort-selection}
\appcontentssubsection{app:task-cohort-coverage}
\appcontentssection{app:protocols}
\appcontentssubsection{app:shared-inference}
\appcontentssubsection{app:geometry-protocol}
\appcontentssubsection{app:table-state-protocol}
\appcontentssubsection{app:gbdi-protocol}
\appcontentssubsection{app:model-configurations}
\appcontentssubsection{app:discovery-metrics}
\appcontentssection{app:additional-results}
\appcontentssubsection{app:geometry-marginals}
\appcontentssubsection{app:joint-geometry}
\appcontentssubsection{app:external-wikisql}
\appcontentssection{app:method-ablations}
\appcontentssubsection{app:equal-call-randomization}
\appcontentssubsection{sec:core-ablations}
\appcontentssubsection{app:skill-prompt-ablations}
\appcontentssubsection{app:discovery-answer-decomposition}
\appcontentssubsection{app:view-budget-support}
\appcontentssubsection{app:view-count-qa}
\appcontentssection{app:uncertainty}
\appcontentssubsection{app:cluster-bootstrap}
\appcontentssubsection{app:interface-attenuation}
\appcontentssubsection{app:table-state-contrasts}
\appcontentssubsection{app:gbdi-discovery-qa}
\appcontentssection{app:prompts}
\appcontentssubsection{app:geometry-prompts}
\appcontentssubsection{app:table-state-prompts}
\appcontentssubsection{app:inspection-schemas}
\appcontentssubsection{app:evidence-processing}
\appcontentssubsection{app:intervention-guidance}
\appcontentssubsection{app:qa-conversation}
\endgroup
\clearpage

%% file: appendices/appendix_a_target_coordinates.tex
\section{Human-Reviewed Reference}
\label{app:target-coordinates}

This section documents the 154-instance reference introduced in
Section~\ref{sec:problem}, including its coverage, review procedure,
and the subsets used in the experiments. The annotations
identify the critical cells $\mathcal C_i$, their containing rows
$\mathcal R_i$, and supported interventions under the definitions in the main text.

\subsection{Reference Coverage}
\label{app:reference-coverage}

The reference comprises 154 non-clean RADAR-T instances from 34 base tasks
and 24 source tables. It contains
1,162 distinct instance--row critical locations. Each retained annotation
identifies the affected row and records the implicated columns,
visible evidence, and an \textsc{Edit} or \textsc{Drop} action; an
\textsc{Edit} also records the supported replacement values.

Of the 154 instances, 141 have nonempty critical-row sets and 13 have empty
sets. The table-state comparison in Section~\ref{sec:intervention-motivation}
uses all 154 instances. CDR for the multi-view analysis in
Section~\ref{sec:end-to-end-validation} uses the 141 nonempty sets; the other
13 instances are excluded because complete discovery would be vacuous.
Their false-positive rates are reported separately in
Appendix~\ref{app:view-budget-support}.

\subsection{Review Procedure}
\label{app:human-review-protocol}

Four paid undergraduate annotators worked in two pairs, reviewing 79 and 75
instances, respectively. Within each pair, both annotators assessed each
instance independently. An additional reviewer, one of the authors, checked
all annotations and adjudicated disagreements.

Reviewers inspected the complete table and question to identify critical cells
and supported interventions according to Section~\ref{sec:problem}.
The gold answer was used only to check the resulting computation, not to
infer error locations or replacement values.

\subsection{Geometry Cohort}
\label{app:geometry-cohort-selection}

The geometry study in Section~\ref{sec:geometry} uses 60 of the reviewed
instances, with 12 from each of the five error types. Selection requires the
answer-preserving row-order assumption and feasible controlled layouts.
The cohort spans 29 base tasks and 21 source tables. All 60 instances have
at least one critical row and enter the position experiment; 59 have multiple
critical rows and also enter the dispersion experiment.

We apply equal quotas across error types and layout feasibility checks using
annotation metadata, without using system outputs. The selected cohort is
fixed across systems and inference modes. For every retained layout, we verify
that the rows moved and scored are exactly $\mathcal R_i$.
Appendix~\ref{app:geometry-protocol} describes the layout construction.

\subsection{Task Coverage and Cohort Membership}
\label{app:task-cohort-coverage}

Table~\ref{tab:task-cohort-membership} lists all 34 base tasks in the
human-reviewed reference and their membership in the position and dispersion
cohorts. Task IDs and source identifiers follow RADAR~\citep{gu2025radar};
Table~\ref{tab:cohort-source-datasets} gives the corresponding source dataset
names. The question summaries identify the tasks; the experiments use the
original RADAR questions.

Each membership cell lists the included error types: \texttt{M} = missing
values, \texttt{B} = bad values, \texttt{O} = outliers, \texttt{F} = formatting
errors, and \texttt{L} = logical inconsistencies. A dash denotes no included
instance. Within these cohorts, each task--error-type pair identifies one
RADAR-T instance. The reviewed column includes all 154 instances, including
the 13 with empty critical-row sets.

The 60-instance position cohort and 59-instance dispersion cohort contain the
same 29 base tasks from 21 source tables. Their only membership difference is
the logical-inconsistency variant of \texttt{olympics-country}: it has one
critical row and is therefore included only in the position experiment.

\input{tables/task_cohort_membership}
\input{tables/cohort_source_datasets}

%% file: tables/task_cohort_membership.tex
\begingroup
\small
\setlength{\tabcolsep}{3pt}
\begin{longtable}{@{}>{\raggedright\arraybackslash}p{0.54\linewidth}cccc@{}}
\caption{Task coverage and exact error-type membership of the three cohorts. Each listed error type identifies one RADAR-T instance.}\label{tab:task-cohort-membership}\\
\toprule
RADAR task ID and question summary & Source & Reviewed & Position & Dispersion \\
\midrule
\endfirsthead
\multicolumn{5}{l}{\tablename~\thetable: Task coverage and cohort membership (continued).}\\
\toprule
RADAR task ID and question summary & Source & Reviewed & Position & Dispersion \\
\midrule
\endhead
\midrule
\multicolumn{5}{r}{\emph{Continued on next page.}}\\
\endfoot
\bottomrule
\endlastfoot
\texttt{actor-age-gaps}\newline Mean age gap for couples with an older male actor. & D12 & \texttt{M B O F L} & \texttt{M O F L} & \texttt{M O F L} \\
\addlinespace[2pt]
\texttt{board-games-min-players}\newline Count games meeting both player-count limits. & D11 & \texttt{M B O F L} & \texttt{M B L} & \texttt{M B L} \\
\addlinespace[2pt]
\texttt{car-co2}\newline Mean estimated car emissions for 2018--2023. & D06 & \texttt{M B O F L} & \texttt{F} & \texttt{F} \\
\addlinespace[2pt]
\texttt{daily-activity}\newline Mean distance share from moderate-or-higher activity. & D20 & \texttt{M B O F L} & \texttt{O F} & \texttt{O F} \\
\addlinespace[2pt]
\texttt{employee-years}\newline Count older employees with at least five years of tenure. & D25 & \texttt{M B O F L} & \textemdash & \textemdash \\
\addlinespace[2pt]
\texttt{england-wales-ethnicity}\newline Find the area ranked seventh by Black Caribbean population. & D14 & \texttt{M B O F L} & \textemdash & \textemdash \\
\addlinespace[2pt]
\texttt{england-wales-housing}\newline Share of lone-parent families in one- or two-bedroom homes. & D15 & \texttt{M B O F L} & \texttt{M B O} & \texttt{M B O} \\
\addlinespace[2pt]
\texttt{england-wales-housing-bedroom-count}\newline Estimate the total bedrooms across households. & D15 & \texttt{M B O F L} & \texttt{B O} & \texttt{B O} \\
\addlinespace[2pt]
\texttt{football-european-league-goal-diff}\newline Mean absolute gap between expected and scored goals for top-five teams. & D26 & \texttt{M B O F L} & \texttt{M} & \texttt{M} \\
\addlinespace[2pt]
\texttt{influenza-like-illness}\newline Median weekly case count for ages 25--64. & D07 & \texttt{M B O F L} & \texttt{M O L} & \texttt{M O L} \\
\addlinespace[2pt]
\texttt{movies-roi}\newline Mean film return relative to production budget. & D08 & \texttt{M B O F L} & \texttt{F} & \texttt{F} \\
\addlinespace[2pt]
\texttt{nba-players-best-shooters}\newline Identify leading players by two shooting-efficiency measures. & D27 & \texttt{M B O F L} & \texttt{F} & \texttt{F} \\
\addlinespace[2pt]
\texttt{northern-hemisphere-eelgrass-habitats}\newline Sum salinity readings at US sampling sites. & D17 & \texttt{M B O F} & \texttt{F} & \texttt{F} \\
\addlinespace[2pt]
\texttt{nurses-hourly-salary}\newline Mean state-level median hourly nurse pay. & D09 & \texttt{M B O F L} & \texttt{M B O L} & \texttt{M B O L} \\
\addlinespace[2pt]
\texttt{nurses-salary-difference}\newline Mean state-level gap between high and median nurse pay. & D09 & \texttt{M B O F} & \texttt{B O} & \texttt{B O} \\
\addlinespace[2pt]
\texttt{nyc-green-taxis}\newline Count trips in hours with the longest mean duration. & D01 & \texttt{M B O F L} & \texttt{B} & \texttt{B} \\
\addlinespace[2pt]
\texttt{nyc-green-taxis-passengers}\newline Sum passenger counts across taxi trips. & D01 & \texttt{M B O F L} & \texttt{O F} & \texttt{O F} \\
\addlinespace[2pt]
\texttt{nyc-green-taxis-rates}\newline Mean total trip charge per mile. & D01 & \texttt{M B O F L} & \texttt{B L} & \texttt{B L} \\
\addlinespace[2pt]
\texttt{nyc-regents-exam-scores}\newline Count passing students in the most frequent school group. & D02 & \texttt{M B O F L} & \texttt{M B L} & \texttt{M B L} \\
\addlinespace[2pt]
\texttt{nyc-regents-exam-scores-2}\newline Compare counts of high-score-dominant and low-score-dominant rows. & D02 & \texttt{F} & \textemdash & \textemdash \\
\addlinespace[2pt]
\texttt{nyc-regents-exam-scores-borough}\newline Count passing students in the most-tested borough. & D02 & \texttt{M B O F L} & \texttt{L} & \texttt{L} \\
\addlinespace[2pt]
\texttt{olympics-country}\newline Find the highest team mean medals per Games. & D13 & \texttt{M B F L} & \texttt{F L} & \texttt{F} \\
\addlinespace[2pt]
\texttt{olympics-gold-winners}\newline Mean athlete age across gold-medal performances. & D13 & \texttt{M B O F L} & \texttt{B O F} & \texttt{B O F} \\
\addlinespace[2pt]
\texttt{pet-respiratory-motion}\newline Mean speed between consecutive recorded timestamps. & D05 & \texttt{M B O F L} & \texttt{L} & \texttt{L} \\
\addlinespace[2pt]
\texttt{physical-health-exam-bmi}\newline Mean BMI for men with no smoking history. & D18 & \texttt{M B O F L} & \texttt{O F L} & \texttt{O F L} \\
\addlinespace[2pt]
\texttt{physical-health-exam-rhr}\newline Compare low-heart-rate counts across two age groups. & D18 & \texttt{O F L} & \textemdash & \textemdash \\
\addlinespace[2pt]
\texttt{queensland-water-emissions}\newline Average coordinates of a frequency-weighted facility centroid. & D03 & \texttt{O L} & \textemdash & \textemdash \\
\addlinespace[2pt]
\texttt{ross-sea-algal-pigment}\newline Find chlorophyll c2 at the median chlorophyll a value. & D16 & \texttt{M F L} & \texttt{M} & \texttt{M} \\
\addlinespace[2pt]
\texttt{sales-2}\newline Sum units ordered during the fourth quarter. & D24 & \texttt{M B F L} & \texttt{B} & \texttt{B} \\
\addlinespace[2pt]
\texttt{traffic-violations}\newline Mean vehicle age in violations recorded during 2014--2024. & D04 & \texttt{M B F L} & \texttt{M B L} & \texttt{M B L} \\
\addlinespace[2pt]
\texttt{traffic-violations-speeding}\newline Mean excess speed for violations over 20 mph above the limit. & D04 & \texttt{M B O F L} & \texttt{M O L} & \texttt{M O L} \\
\addlinespace[2pt]
\texttt{udemy-classes-price}\newline Count high-priced courses discounted by more than 88 percent. & D23 & \texttt{M B O F L} & \texttt{M} & \texttt{M} \\
\addlinespace[2pt]
\texttt{udemy-classes-rating}\newline Count courses with a rating above 4.1. & D23 & \texttt{M B O F L} & \texttt{M F} & \texttt{M F} \\
\addlinespace[2pt]
\texttt{ultra-trail-races-rank}\newline Mean age among runners finishing in the top five. & D10 & \texttt{M B O F L} & \texttt{B O F} & \texttt{B O F} \\
\addlinespace[2pt]
\midrule
\textbf{Instances} & & \textbf{154} & \textbf{60} & \textbf{59} \\
Distinct base tasks & & 34 & 29 & 29 \\
Distinct source tables & & 24 & 21 & 21 \\
\end{longtable}
\endgroup

%% file: tables/cohort_source_datasets.tex
\begin{table}[H]
\centering\small
\caption{Source datasets represented in the human-reviewed reference. Source identifiers and dataset names follow RADAR.}\label{tab:cohort-source-datasets}
\begin{tabularx}{\linewidth}{@{}lX@{}}
\toprule
Source & Dataset \\
\midrule
D01 & 2021 Green Taxi Trip Data \\
D02 & 2014--15 to 2017--19 NYC Regents Exam Results -- Public \\
D03 & Emissions from Industrial Facilities in Queensland -- 2004 \\
D04 & Traffic Violations \\
D05 & Tracking data, Subject (a) MC Motion \\
D06 & Fuel Economy Data \\
D07 & Outpatient Illness and Viral Surveillance \\
D08 & Movies Bechdel Test \\
D09 & Registered Nurses \\
D10 & Ultra Trail Running \\
D11 & Board Games \\
D12 & Hollywood Age Gaps \\
D13 & Olympics Athletes and Medals \\
D14 & Ethnic group (England and Wales) 2011 \\
D15 & Household Composition by Number of bedrooms 2011 \\
D16 & Algal Pigment Concentrations in Ross Sea \\
D17 & Eelgrass Biomass and Diversity (ZEN) \\
D18 & Framingham Heart Study Dataset \\
D20 & FitBit Fitness Tracker Data \\
D23 & Udemy Finance Courses \\
D24 & Sales Data \\
D25 & IBM HR Analytics \\
D26 & Football Expected Goals \\
D27 & 2023--2024 NBA Player Stats \\
\bottomrule
\end{tabularx}
\end{table}

%% file: appendices/appendix_b_protocols.tex
\section{Experimental Protocols}
\label{app:protocols}

The protocols below follow the geometry, table-state, and GBDI studies in
Sections~\ref{sec:geometry}--\ref{sec:end-to-end-validation}.
Appendix~\ref{app:target-coordinates} defines the reviewed evaluation subsets;
Appendix~\ref{app:prompts} provides the exact prompts and output schemas.

\subsection{Benchmark and Shared Inference Conditions}
\label{app:shared-inference}

RADAR-T contains 313 instances from 53 base tasks and 27 source
tables~\citep{gu2025radar}. A base task specifies a question and source data;
instances apply artifact variants at the table-size setting for the split.
The records use \texttt{example\_id} for instances and \texttt{task\_id} for
base tasks. Constructed views are repeated layouts of an instance, not
independent tasks. The full benchmark is used for end-to-end QA; the reviewed
reference and its geometry subsets are described in
Appendix~\ref{app:target-coordinates}.

At inference time, each system receives the question, displayed table,
condition-specific instructions, and any evidence generated within that
condition. Human annotations are used for offline scoring and the controlled
table-state study, where errors are marked or repaired before inference.
No condition receives the gold answer.

\subsection{Geometry Diagnostic}
\label{app:geometry-protocol}

We adapt the Direct and Code Agent interfaces from RADAR~\citep{gu2025radar}
to critical-row discovery. Under Direct, systems read the complete table view and return candidate row IDs in
one response. Code Agent systems receive the same view and objective and may
use up to five responses, including the final candidate submission, with
read-only Python inspection. The visible dataframe is restored after each
inspection command, and the task terminates at localization rather than final
QA. Here, localization is the runtime term for critical-row discovery.

Every view preserves row values and multiplicities. The question--table prompt
arrangement is fixed within each comparison. We call a fixed order of the
remaining, noncritical rows a background and reuse it across the compared layouts.

\paragraph{Position.}

The position experiment uses the 60-instance cohort in
Appendix~\ref{app:geometry-cohort-selection}. For each instance
and each of three backgrounds formed by the remaining rows, the same
critical-row block is placed so that its realized $\mu$ falls in one of five equal-width bands.
The feasible block location is randomized within the assigned band,
subject to preserving critical-row identities, critical-row count, row contents, question,
and prompt template. Each system--mode pair therefore evaluates
$60\times3\times5=900$ views. Complete discovery rate is reported separately
for each of the five position bands.

\paragraph{Dispersion.}

The dispersion cohort excludes the one instance with a single critical row,
leaving 59 instances with at least two critical rows. For
each instance, three target mean positions $\mu$ (25\%, 50\%, and 75\%) and three
backgrounds are crossed with Compact, Medium, and Wide layouts. Critical-row
identities, values, count, $\mu$, question, and the relative order of the
remaining rows are held fixed within each comparison triplet. The construction
yields $59\times3\times3\times3=1{,}593$ views per system--mode pair. Compact uses
the minimum consecutive critical-row span; Medium uses an intermediate symmetric
span; Wide uses the largest feasible symmetric span within a normalized radius
of 0.2. The audit records realized $\mu$ and $\sigma$ for every constructed
view.

\subsection{Table-State Diagnostic}
\label{app:table-state-protocol}

The diagnostic in Section~\ref{sec:intervention-motivation} compares Raw,
Discovery, and Action on the 154 instances in
Appendix~\ref{app:target-coordinates}, under both Direct and Code Agent. Raw uses the canonical table. Discovery
wraps each reviewed critical cell value in \texttt{<<ERROR:...>>} while
retaining the original value; it does not provide a replacement or a separate
evidence packet. Action starts from the canonical table, applies the reviewed
replacement values for \textsc{Edit}, and removes rows assigned \textsc{Drop}.
The Discovery instruction identifies marked cells as verified errors. The
Action instruction states that the reviewed errors relevant to the answer have
been repaired. Both notices precede the usual user turn; the system prompt is
unchanged within each inference mode. Raw Direct uses the released RADAR
Direct prompt. Conditions within a mode share the answer scorer and response
allowance. Requests for all three systems and both modes in this diagnostic omit
an explicit output-token cap; provider defaults still apply. Comparisons are paired
by instance and clustered by source table for uncertainty estimation.

\subsection{GBDI and Matched Controls}
\label{app:gbdi-protocol}

GBDI uses the $M$ independently permuted row orders in
Section~\ref{sec:gbdi-workflow}. We denote these $M$ independent inspections
by Random-$M$. Repeated-$M$ is the call-matched control: it presents the
unchanged released row order in $M$ independent inspection contexts.
Inspection contexts use the same decoding parameters within a system.
Nominations are mapped to the original rows; every valid nomination
and its independent-view support count are retained. The final Code Agent receives the canonical table, question,
evidence ledger, and Intervention Skill. The intervention turn checks visible
evidence and prepares a persistent \texttt{work\_df}; the same conversation
and Python namespace continue to final QA. The primary configuration is
Random-5.
Visible evidence, rather than nomination frequency alone, determines whether an
intervention is supported.

The Qwen, GLM, and MiniMax benchmark collections use the same instance--view
permutations, and every view retains all rows. DeepSeek-V4.1-Flash also uses five
independently permuted views. Permutation construction
does not use human annotations.
The union ledger retains every valid nomination. A support threshold of at
least two is examined separately in the ledger-support sensitivity analysis;
end-to-end QA uses the union ledger.

Within a system, Random-$M$ and Repeated-$M$ share the call allowance,
discovery prompt, output schema, parser, and final QA
protocol. The guidance factorial crosses the two view conditions with
Generic/Skill on the same 313 instances for Qwen, GLM, and MiniMax; GPT and
DeepSeek-V4.1-Flash appear only in the
complete-system evaluation. Generic specifies general operation boundaries;
Skill adds artifact-specific rules and parsing, validation, and submission
guidance. Within each layout, both guidance conditions
receive the same ledger, so only the guidance text changes. The reserved turn,
persistent namespace, and five-response final QA allowance remain fixed.

The blank-ledger diagnostic preserves the canonical table, question, Skill,
intervention turn, and final budget while removing ledger coordinates and
observations. It measures the contribution of ledger availability within the
candidate-guided workflow.

\subsection{System Configurations}
\label{app:model-configurations}

Table~\ref{tab:model-provenance} lists the systems evaluated on RADAR-T and their request
identifiers. The benchmark and component analyses use the same recorded
GBDI outcomes (Random-5+Skill). We conducted the API evaluations during
August and September 2026.

The Direct baseline uses the unchanged RADAR prompt, canonical RADAR table,
original question, and required answer parser on all 313 instances.

\begin{table}[H]
    \centering\small
    \setlength{\tabcolsep}{5pt}
    \caption{Systems evaluated on RADAR-T and their API providers. API identifiers are request
    identifiers rather than checkpoint hashes.}
    \label{tab:model-provenance}
    \begin{tabular}{@{}lll@{}}
        \toprule
        System & API identifier & Provider \\
        \midrule
        Qwen Plus & \texttt{qwen-plus} & Bailian \\
        MiniMax-M2.5 & \texttt{MiniMax-M2.5} & Bailian \\
        GPT-5 Mini & \texttt{openai/gpt-5-mini} & OpenRouter/OpenAI \\
        DeepSeek-V4.1-Flash & \texttt{deepseek-flash} & DeepSeek native API \\
        GLM-4.5-Air & \texttt{glm-4.5-air} & Bailian \\
        \bottomrule
    \end{tabular}
\end{table}

The Bailian requests specify temperature 0 and top-$p$ 1. DeepSeek-V4.1-Flash uses
temperature 0 with thinking disabled and no submitted response-token cap.
GPT-5 Mini omits these sampling parameters. For its GBDI runs, requests specify
low reasoning effort, seed 408, a 12,000-token response cap, and disabled provider
fallback. Within each system, the prompt, parser, and renderer are shared
across view families. The Random-5 and Repeated-5 configurations use five
inspection contexts, one intervention turn, and at most five final QA responses.

\subsection{Scoring and Reporting}
\label{app:discovery-metrics}

The Direct baseline extracts the answer following the required
\texttt{The answer is:} prefix. Format retries regenerate a response, count
against the three-attempt allowance, and contribute to observed usage when
the provider returns usage information. The Code Agent baseline, GBDI, and the view-family controls require a
valid \texttt{done} submission whose \texttt{kwargs.answer} field contains the
answer. The extraction rule is shared within each submission interface, while
Direct and Code retain their respective output formats. All reported runs were
rescored with these rules. RADAR-T discovery predictions are matched to the canonical
critical-row sets in Appendix~\ref{app:target-coordinates}. Invalid formats,
budget exhaustion, tool errors, and transport failures are recorded separately;
all count as empty predictions or incorrect answers in the primary denominator.

CDR measures complete coverage without penalizing additional nominations. We
therefore also report row-level precision, recall, and F1. For an evaluated
view or ledger, let $\widehat{\mathcal R}$ be the set of nominated row IDs and
$\mathcal R$ the corresponding nonempty reference critical-row set. Then
\begin{equation}
\operatorname{Prec}=\frac{|\widehat{\mathcal R}\cap\mathcal R|}
{|\widehat{\mathcal R}|},
\qquad
\operatorname{Rec}=\frac{|\widehat{\mathcal R}\cap\mathcal R|}
{|\mathcal R|},
\qquad
\operatorname{F1}=\frac{2|\widehat{\mathcal R}\cap\mathcal R|}
{|\widehat{\mathcal R}|+|\mathcal R|}.
\label{eq:discovery-recall-precision}
\end{equation}
Precision is zero for an empty prediction. For parsed geometry Direct
submissions, unmatched row IDs remain false-positive nominations.
Each metric is computed before
macro averaging; F1 is not the harmonic mean of the reported average
precision and recall. Geometry metrics average over all views in the stated
condition, with equal numbers of views per instance. Multi-view metrics
average over the 141 reviewed instances with nonempty critical-row sets;
the budget analysis additionally averages over the specified view subsets.
Mean rows is the average nomination count over all 154 reviewed instances.
For the 13 empty-reference instances, Empty FP is the fraction with at least
one nominated row. All rate columns are percentages. These discovery metrics
use the same reference and predictions as the corresponding CDR.

Discovery contrasts are paired instance-level CDR differences averaged over
the construction backgrounds and, for dispersion, the same values of $\mu$.
The dispersion effect is $\operatorname{CDR}(\text{Wide})-
\operatorname{CDR}(\text{Compact})$. Mode differences subtract the Direct
effect from the Code Agent effect. All calculations are performed before rounding.
Binary QA results are reported as benchmark QA accuracy on the stated evaluation population.
Appendix~\ref{app:uncertainty} specifies the paired contrasts and cluster
bootstrap procedures.

%% file: appendices/appendix_c_additional_results.tex
\section{Additional Geometry Results}
\label{app:additional-results}

This section supplements the geometry study in Section~\ref{sec:geometry}
with the complete marginal results, joint position--dispersion breakdowns,
and an external validation on WikiSQL.
The RADAR-T cohort and layout construction are specified in
Appendices~\ref{app:geometry-cohort-selection} and~\ref{app:geometry-protocol};
paired uncertainty estimates appear in Appendix~\ref{app:uncertainty}.

\subsection{Position and Dispersion Results}
\label{app:geometry-marginals}

Table~\ref{tab:geometry-interface-full} reports the complete discovery rate
(CDR) values plotted in Figure~\ref{fig:geometry-main}, together with macro
precision, recall, and F1 as defined in Appendix~\ref{app:discovery-metrics}. The table has two
parts: (a) the five position bands and (b) the three dispersion levels,
averaged over the same three values of $\mu$.
The position experiment uses 60 instances and 900 views per system--mode pair;
the dispersion experiment uses 59 instances with multiple critical rows and
1,593 views per pair. Both experiments use
three fixed backgrounds, and invalid submissions count as discovery failures.

\input{tables/geometry_interface_three_model}

\subsection{Joint Position-by-Dispersion Results}
\label{app:joint-geometry}

The dispersion experiment contains nine layouts per instance and background.
Table~\ref{tab:joint-geometry-layout} reports their realized coordinates. The
mean nearest-neighbor dispersion is 0.0081, 0.0677, and 0.1268 for Compact,
Medium, and Wide, respectively, at each value of $\mu$.
Within each instance--background pair, $\mu$ is fixed across dispersion levels
and $\sigma$ is fixed across positions. All systems and both inference modes use
the same layouts.

\input{tables/human_joint_geometry_layout}

Tables~\ref{tab:joint-geometry-direct} and~\ref{tab:joint-geometry-code}
report CDR, precision, recall, and F1 for every position--dispersion combination. Every condition uses the
same 59 instances and three backgrounds, yielding 177 views. The three values of
$\mu$ are distinct from the five bands in the contiguous critical-row position
experiment. Within each position group, $\sigma$ varies while $\mu$ remains fixed. Invalid
submissions count as discovery failures.

\input{tables/human_joint_geometry_direct}

Under Direct, CDR decreases from Compact to Wide at each $\mu$ for
every system. Intermediate values show local variation; for MiniMax at
$\mu=75\%$, CDR is 51.4\% for Medium and 55.4\% for Wide.

\input{tables/human_joint_geometry_code}

Code Agent results also vary across the tested mean positions.

\subsection{Error Discovery on WikiSQL}
\label{app:external-wikisql}

We constructed missingness instances from 41 WikiSQL tables~\citep{zhong2017seq2sqlgeneratingstructuredqueries},
following the RADAR missingness setting~\citep{gu2025radar}.
Each question selects rows by a rank threshold and asks for the sum of
another numeric column. We replace three question-relevant values in the
rank column with empty cells. We evaluate Direct discovery
with Qwen-Turbo and Qwen3-8B using the position and dispersion manipulations
in Section~\ref{sec:geometry}. Position uses q1--q5 and three fixed backgrounds;
dispersion crosses Compact, Medium, and Wide with three mean positions
and one fixed background. We score all planned views against the injected
critical rows using the metrics in Appendix~\ref{app:discovery-metrics},
with invalid submissions treated as empty predictions.

\input{figures/external_wikisql_geometry}
\input{tables/external_wikisql_geometry}

Figure~\ref{fig:external-wikisql} and Table~\ref{tab:external-wikisql}
show a decrease in CDR from Compact to Wide layouts: from 80.5\% to 48.0\%
for Qwen-Turbo and from 64.2\% to 35.0\% for Qwen3-8B.
Position effects are less uniform. CDR is higher at q5 than q1 for both
systems, but neither curve is monotonic; Qwen3-8B recall is slightly lower
at q5. In this external missingness setting, the dispersion contrast is
more consistent across the two systems than the position contrast.

%% file: tables/geometry_interface_three_model.tex
\begin{table}[H]
\centering\small
\setlength{\tabcolsep}{3pt}\renewcommand{\arraystretch}{1.06}
\caption{Discovery metrics by critical-row position and dispersion. Precision
(Prec.), recall (Rec.), and F1 are macro averages; all entries are percentages.
Each position band uses 180 views of 60 instances. Each dispersion level uses
531 views of 59 instances, averaged over three values of $\mu$.}
\label{tab:geometry-interface-full}
\begin{tabular}{@{}llrrrrrrrr@{}}\toprule
\multicolumn{10}{c}{\textbf{(a) Position $\mu$}} \\
\midrule
 & & \multicolumn{4}{c}{Direct} & \multicolumn{4}{c}{Code Agent} \\
\cmidrule(lr){3-6}\cmidrule(l){7-10}
System & Layout & CDR & Prec. & Rec. & F1 & CDR & Prec. & Rec. & F1 \\
\midrule
\multirow{5}{*}{Qwen Plus} & q1 & 61.1 & 70.6 & 76.8 & 70.2 & 66.1 & 68.8 & 72.4 & 67.5 \\
 & q2 & 68.9 & 72.4 & 77.8 & 71.9 & 75.6 & 75.7 & 80.1 & 74.8 \\
 & q3 & 66.7 & 71.7 & 76.0 & 70.9 & 76.1 & 76.1 & 82.2 & 76.9 \\
 & q4 & 70.6 & 75.5 & 79.5 & 74.7 & 73.9 & 75.4 & 79.2 & 75.1 \\
 & q5 & 72.8 & 76.1 & 82.0 & 75.7 & 73.3 & 75.4 & 78.6 & 73.9 \\
\addlinespace[0.35em]
\multirow{5}{*}{MiniMax-M2.5} & q1 & 47.8 & 59.0 & 59.4 & 56.6 & 38.3 & 44.8 & 42.3 & 42.6 \\
 & q2 & 60.6 & 65.5 & 69.0 & 65.0 & 38.3 & 44.1 & 42.9 & 42.7 \\
 & q3 & 58.9 & 68.5 & 67.7 & 66.2 & 48.9 & 51.6 & 52.1 & 51.1 \\
 & q4 & 56.7 & 67.5 & 67.6 & 65.8 & 43.3 & 46.8 & 45.9 & 45.5 \\
 & q5 & 59.4 & 68.5 & 68.0 & 66.2 & 43.3 & 50.0 & 47.7 & 47.8 \\
\addlinespace[0.35em]
\multirow{5}{*}{GPT-5 Mini} & q1 & 43.9 & 59.6 & 56.0 & 55.8 & 68.9 & 75.3 & 74.1 & 72.6 \\
 & q2 & 52.8 & 65.4 & 62.4 & 61.7 & 73.9 & 79.6 & 79.7 & 77.2 \\
 & q3 & 56.1 & 61.7 & 61.2 & 60.1 & 75.0 & 77.7 & 78.7 & 76.3 \\
 & q4 & 60.0 & 69.3 & 66.9 & 65.9 & 72.2 & 75.4 & 77.2 & 73.8 \\
 & q5 & 60.6 & 66.7 & 66.9 & 64.9 & 72.2 & 77.1 & 77.6 & 75.5 \\
\bottomrule
\end{tabular}

\vspace{0.5em}
\begin{tabular}{@{}llrrrrrrrr@{}}\toprule
\multicolumn{10}{c}{\textbf{(b) Dispersion $\sigma$}} \\
\midrule
 & & \multicolumn{4}{c}{Direct} & \multicolumn{4}{c}{Code Agent} \\
\cmidrule(lr){3-6}\cmidrule(l){7-10}
System & Layout & CDR & Prec. & Rec. & F1 & CDR & Prec. & Rec. & F1 \\
\midrule
\multirow{3}{*}{Qwen Plus} & Compact & 72.3 & 75.7 & 81.8 & 75.8 & 72.1 & 74.6 & 78.4 & 73.8 \\
 & Medium & 59.5 & 75.8 & 78.9 & 74.1 & 70.6 & 74.9 & 78.9 & 74.3 \\
 & Wide & 56.1 & 76.4 & 77.0 & 73.1 & 69.9 & 75.5 & 78.6 & 74.5 \\
\addlinespace[0.35em]
\multirow{3}{*}{MiniMax-M2.5} & Compact & 59.3 & 67.2 & 68.9 & 65.6 & 44.4 & 50.8 & 48.8 & 48.7 \\
 & Medium & 49.7 & 66.8 & 65.2 & 63.5 & 38.0 & 45.6 & 42.4 & 42.8 \\
 & Wide & 49.7 & 63.6 & 63.8 & 61.2 & 40.7 & 48.1 & 45.6 & 45.7 \\
\addlinespace[0.35em]
\multirow{3}{*}{GPT-5 Mini} & Compact & 54.2 & 64.7 & 62.0 & 61.3 & 70.2 & 76.9 & 76.8 & 74.8 \\
 & Medium & 51.4 & 64.8 & 61.4 & 61.1 & 68.9 & 78.4 & 76.4 & 75.1 \\
 & Wide & 46.3 & 62.8 & 59.4 & 58.9 & 68.2 & 78.4 & 75.6 & 74.8 \\
\bottomrule
\end{tabular}
\end{table}

%% file: tables/human_joint_geometry_layout.tex
\begin{table}[H]
\centering\small
\setlength{\tabcolsep}{12pt}\renewcommand{\arraystretch}{1.12}
\caption{Realized geometry of the $3\times3$ layout design. Each cell gives mean position $\bar\mu$ / mean dispersion $\bar\sigma$.}\label{tab:joint-geometry-layout}
\begin{tabular}{@{}lrrr@{}}\toprule
Target $\mu$ & Compact & Medium & Wide \\
\midrule
25\% & 0.2495 / 0.0081 & 0.2495 / 0.0677 & 0.2495 / 0.1268 \\
50\% & 0.5008 / 0.0081 & 0.5008 / 0.0677 & 0.5008 / 0.1268 \\
75\% & 0.7499 / 0.0081 & 0.7499 / 0.0677 & 0.7499 / 0.1268 \\
\bottomrule
\end{tabular}
\end{table}

%% file: tables/human_joint_geometry_direct.tex
\begin{table}[H]
\centering\small
\setlength{\tabcolsep}{6pt}\renewcommand{\arraystretch}{1.06}
\caption{Joint position--dispersion discovery metrics under Direct.
Each condition uses the same 59 instances and three backgrounds (177 views).
Precision, recall, and F1 are macro averages; all entries are percentages.}
\label{tab:joint-geometry-direct}
\begin{tabular}{@{}lclrrrr@{}}\toprule
System & $\mu$ & Dispersion & CDR & Prec. & Rec. & F1 \\
\midrule
Qwen Plus & \multirow{3}{*}{25\%} & Compact & 71.2 & 72.7 & 80.2 & 73.3 \\
 &  & Medium & 53.1 & 71.9 & 75.8 & 70.5 \\
 &  & Wide & 52.0 & 73.0 & 74.0 & 70.0 \\
\addlinespace[0.25em]
 & \multirow{3}{*}{50\%} & Compact & 70.1 & 75.9 & 81.0 & 75.7 \\
 &  & Medium & 59.3 & 77.2 & 79.1 & 74.9 \\
 &  & Wide & 54.2 & 76.9 & 77.6 & 73.8 \\
\addlinespace[0.25em]
 & \multirow{3}{*}{75\%} & Compact & 75.7 & 78.3 & 84.3 & 78.4 \\
 &  & Medium & 66.1 & 78.2 & 81.8 & 76.9 \\
 &  & Wide & 62.1 & 79.4 & 79.3 & 75.5 \\
\midrule
MiniMax-M2.5 & \multirow{3}{*}{25\%} & Compact & 57.1 & 66.1 & 67.8 & 63.7 \\
 &  & Medium & 48.6 & 67.7 & 64.5 & 63.7 \\
 &  & Wide & 45.2 & 61.4 & 61.2 & 58.4 \\
\addlinespace[0.25em]
 & \multirow{3}{*}{50\%} & Compact & 61.0 & 67.1 & 69.0 & 66.0 \\
 &  & Medium & 49.2 & 63.5 & 63.9 & 61.3 \\
 &  & Wide & 48.6 & 62.4 & 62.9 & 60.9 \\
\addlinespace[0.25em]
 & \multirow{3}{*}{75\%} & Compact & 59.9 & 68.6 & 69.8 & 67.1 \\
 &  & Medium & 51.4 & 69.2 & 67.3 & 65.5 \\
 &  & Wide & 55.4 & 66.9 & 67.4 & 64.4 \\
\midrule
GPT-5 Mini & \multirow{3}{*}{25\%} & Compact & 53.1 & 64.7 & 62.2 & 61.3 \\
 &  & Medium & 47.5 & 61.4 & 57.5 & 57.2 \\
 &  & Wide & 46.9 & 63.7 & 59.0 & 58.7 \\
\addlinespace[0.25em]
 & \multirow{3}{*}{50\%} & Compact & 55.9 & 63.4 & 62.1 & 61.1 \\
 &  & Medium & 53.7 & 64.5 & 62.3 & 61.6 \\
 &  & Wide & 42.9 & 56.6 & 55.0 & 54.0 \\
\addlinespace[0.25em]
 & \multirow{3}{*}{75\%} & Compact & 53.7 & 66.1 & 61.7 & 61.5 \\
 &  & Medium & 53.1 & 68.4 & 64.4 & 64.6 \\
 &  & Wide & 49.2 & 68.1 & 64.0 & 64.1 \\
\bottomrule
\end{tabular}
\end{table}

%% file: tables/human_joint_geometry_code.tex
\begin{table}[H]
\centering\small
\setlength{\tabcolsep}{6pt}\renewcommand{\arraystretch}{1.06}
\caption{Joint position--dispersion discovery metrics under Code Agent.
Each condition uses the same 59 instances and three backgrounds (177 views).
Precision, recall, and F1 are macro averages; all entries are percentages.}
\label{tab:joint-geometry-code}
\begin{tabular}{@{}lclrrrr@{}}\toprule
System & $\mu$ & Dispersion & CDR & Prec. & Rec. & F1 \\
\midrule
Qwen Plus & \multirow{3}{*}{25\%} & Compact & 68.9 & 72.4 & 76.0 & 71.3 \\
 &  & Medium & 70.6 & 72.4 & 76.7 & 72.0 \\
 &  & Wide & 68.9 & 73.2 & 76.4 & 71.8 \\
\addlinespace[0.25em]
 & \multirow{3}{*}{50\%} & Compact & 71.8 & 75.6 & 78.3 & 74.8 \\
 &  & Medium & 68.4 & 78.1 & 80.5 & 77.2 \\
 &  & Wide & 72.9 & 76.3 & 80.4 & 75.9 \\
\addlinespace[0.25em]
 & \multirow{3}{*}{75\%} & Compact & 75.7 & 75.8 & 80.9 & 75.2 \\
 &  & Medium & 72.9 & 74.1 & 79.4 & 73.6 \\
 &  & Wide & 67.8 & 76.9 & 79.0 & 75.6 \\
\midrule
MiniMax-M2.5 & \multirow{3}{*}{25\%} & Compact & 44.6 & 48.6 & 47.9 & 47.0 \\
 &  & Medium & 36.7 & 43.1 & 41.6 & 41.0 \\
 &  & Wide & 37.3 & 44.1 & 42.3 & 42.3 \\
\addlinespace[0.25em]
 & \multirow{3}{*}{50\%} & Compact & 42.9 & 47.9 & 47.1 & 46.7 \\
 &  & Medium & 40.7 & 46.6 & 43.9 & 44.4 \\
 &  & Wide & 43.5 & 50.1 & 47.6 & 47.6 \\
\addlinespace[0.25em]
 & \multirow{3}{*}{75\%} & Compact & 45.8 & 55.9 & 51.4 & 52.4 \\
 &  & Medium & 36.7 & 46.9 & 41.6 & 42.9 \\
 &  & Wide & 41.2 & 50.2 & 46.9 & 47.4 \\
\midrule
GPT-5 Mini & \multirow{3}{*}{25\%} & Compact & 69.5 & 76.4 & 76.8 & 74.6 \\
 &  & Medium & 69.5 & 77.7 & 76.3 & 74.7 \\
 &  & Wide & 66.7 & 78.0 & 74.9 & 74.1 \\
\addlinespace[0.25em]
 & \multirow{3}{*}{50\%} & Compact & 71.2 & 78.5 & 79.0 & 76.2 \\
 &  & Medium & 66.7 & 79.4 & 74.9 & 74.7 \\
 &  & Wide & 70.1 & 78.0 & 75.7 & 74.6 \\
\addlinespace[0.25em]
 & \multirow{3}{*}{75\%} & Compact & 70.1 & 75.8 & 74.7 & 73.4 \\
 &  & Medium & 70.6 & 78.2 & 78.0 & 76.0 \\
 &  & Wide & 67.8 & 79.2 & 76.1 & 75.6 \\
\bottomrule
\end{tabular}
\end{table}

%% file: figures/external_wikisql_geometry.tex
\begin{figure}[H]
\centering
\includegraphics[width=\linewidth]{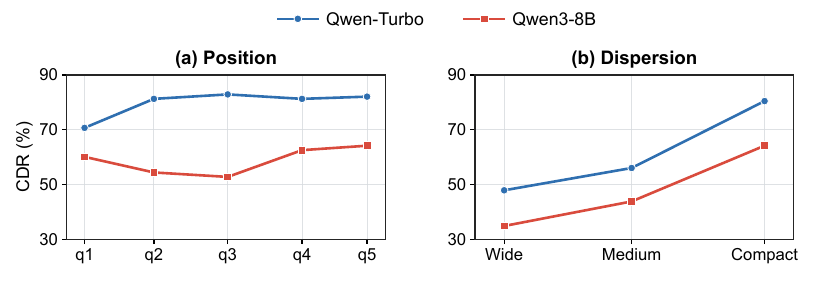}
\caption{Direct discovery on WikiSQL missingness instances. Position runs
from q1 (front) to q5 (back); dispersion runs from Wide to Compact.
Each point averages 123 views of the same 41 instances.}
\label{fig:external-wikisql}
\end{figure}

%% file: tables/external_wikisql_geometry.tex
\begin{table}[H]
\centering\small
\setlength{\tabcolsep}{5pt}\renewcommand{\arraystretch}{1.04}
\caption{Discovery metrics on WikiSQL missingness instances. All entries
are percentages; precision, recall, and F1 are macro averages.
Each system uses 615 position views and 369 dispersion views.}
\label{tab:external-wikisql}
\begin{tabular}{@{}lrrrrrrrr@{}}\toprule
 & \multicolumn{4}{c}{Qwen-Turbo} & \multicolumn{4}{c}{Qwen3-8B} \\
\cmidrule(lr){2-5}\cmidrule(l){6-9}
Layout & CDR & Prec. & Rec. & F1 & CDR & Prec. & Rec. & F1 \\
\midrule
\multicolumn{9}{c}{(a) Position} \\
\addlinespace[0.15em]
q1 & 70.7 & 47.2 & 74.0 & 49.5 & 60.2 & 20.9 & 66.1 & 26.0 \\
q2 & 81.3 & 74.6 & 83.2 & 75.7 & 54.5 & 38.9 & 58.8 & 42.1 \\
q3 & 82.9 & 77.7 & 84.0 & 78.4 & 52.8 & 46.1 & 58.3 & 48.4 \\
q4 & 81.3 & 75.8 & 83.5 & 76.8 & 62.6 & 57.1 & 66.1 & 58.9 \\
q5 & 82.1 & 74.8 & 82.7 & 76.6 & 64.2 & 57.6 & 64.8 & 59.5 \\
\midrule
\multicolumn{9}{c}{(b) Dispersion} \\
\addlinespace[0.15em]
Compact & 80.5 & 72.9 & 82.1 & 74.0 & 64.2 & 56.6 & 67.5 & 58.6 \\
Medium & 56.1 & 78.3 & 78.3 & 73.8 & 43.9 & 50.4 & 61.8 & 51.5 \\
Wide & 48.0 & 80.5 & 73.7 & 74.2 & 35.0 & 49.8 & 58.3 & 48.1 \\
\bottomrule
\end{tabular}
\end{table}

%% file: appendices/appendix_f_method_ablations.tex
\section{Additional GBDI Analyses}
\label{app:method-ablations}
\raggedbottom

The analyses below expand Section~\ref{sec:setup}: we first test view
randomization under equal call budgets, then examine intervention guidance,
the relation between discovery and answer recovery, and sensitivity to view
count and ledger support. They use the protocols in
Appendix~\ref{app:gbdi-protocol} and the reviewed reference in
Appendix~\ref{app:target-coordinates}.

\subsection{Equal-Call View Randomization}
\label{app:equal-call-randomization}

CDR and macro precision, recall, and F1 are evaluated on the 141 reviewed
instances with nonempty critical-row sets. Mean rows and Empty FP follow
Appendix~\ref{app:discovery-metrics}. Within each system, Repeated-5 and Random-5 use the same five-call
allowance, prompt, parser, coordinate mapping, and union rule.

\input{tables/call_matched_discovery}

Random-5 yields higher CDR point estimates for all three systems. Paired
intervals exclude zero for Qwen and MiniMax, but not GLM
(Table~\ref{tab:discovery-uncertainty}). The final QA ledger retains all valid
nominations, which the agent must verify against visible table evidence.
Random-5 also raises recall for all three systems, but precision and F1 do not
improve uniformly: for GLM, they change from 60.5\% to 59.7\% and from 62.5\%
to 62.3\%, respectively. The mean ledger size increases for all three systems.

\subsection{View and Guidance Controls}
\label{sec:core-ablations}

The Qwen Plus factorial analysis crosses Random-5 and Repeated-5 with
Generic and Skill guidance on all 313 RADAR-T instances. Both view conditions
share the discovery prompt, five-call allowance, aggregation rule, and final
QA runtime. Within each view condition, Generic and Skill receive
identical ledgers, tables, questions, and Code Agent response allowances.
Their discovery metrics are therefore shared and reported in
Table~\ref{tab:call-matched-discovery-crossmodel}.
In Table~\ref{tab:core-method-ablations}(a), each guidance cell reports the number of correct answers among
all 313 instances, followed by accuracy. CDR in panel (b) is computed on the
141 reviewed instances with
nonempty critical-row sets. We define
$\Delta_{\mathrm{Skill}}=\text{Skill}-\text{Generic}$ and
$\Delta_{\mathrm{view}}=\text{Random-5}-\text{Repeated-5}$ using unrounded
rates.

\input{tables/core_method_ablations}

With the same five-call allowance, Random-5 raises CDR from 58.9\% to 78.7\%.
Its estimated QA accuracy is 4.2 points higher than that of Repeated-5,
although the paired interval includes zero. Relative to Repeated-5, Random-5
corrects 39 previously incorrect answers and changes 26 correct answers to
incorrect answers, yielding a net gain of 13/313.

With each ledger held fixed, Skill raises QA accuracy by 5.8 points under
Repeated-5 and 6.4 points under Random-5; the latter interval excludes zero.
The blank-ledger diagnostic uses the same Skill-guided QA protocol (180/313).
Relative to this diagnostic, Repeated-5 yields 10 additional correct answers
and Random-5 yields 23. This comparison measures the effect of ledger
availability within the candidate-guided workflow.

\input{tables/core_ablation_artifacts}

With Skill fixed, Random has its largest advantage over Repeated on Logic
(28 versus 17 correct). Skill provides its largest gain over Generic on
Formatting in both layouts: 11 additional correct answers with Repeated and
10 with Random. Under Random, Skill also yields five additional correct
answers each on Logic and Missingness and three on Outliers, while yielding
three fewer on Bad values and leaving Clean unchanged. The aggregate gain is
concentrated in Logic, Formatting, and Missingness.

The factorial experiment compares detailed guidance with generic operation
boundaries under the same intervention-turn budget. The Skill combines
artifact-specific action rules with parsing, validation, and submission
guidance, so this contrast evaluates the complete guidance text.

Under Random-5, the Skill corrects 47 answers and changes 27
correct answers to incorrect answers, yielding a net gain of 6.39
points. The view-by-guidance
interaction is 0.64 points (95\% CI $[-7.92,9.20]$); the Skill gain alone does
not establish an interaction.

The Skill-gain interval also excludes zero under base-task clustering
($[0.96,11.65]$).

\subsection{Intervention Skill Rules}
\label{app:skill-prompt-ablations}

We further remove three instruction groups from the Qwen Plus Skill, one at a
time, while keeping the Random-5 ledger, 313 instances, and final QA protocol
fixed. The recovery variant omits explicit procedures for deriving missing or
invalid values from table evidence. The formatting variant omits rules for
parsing inconsistent formats. The execution variant omits the multi-turn
checks in the Skill; the intervention turn required in every condition
remains in place. These variants delete instructions rather than paraphrasing
the same guidance.

\input{tables/skill_prompt_ablations}

Removing formatting rules reduces overall accuracy from 64.9\% to 60.7\%,
with the paired interval for the change below zero. On Formatting instances,
correct answers fall from 38/53 to 33/53. Removing recovery rules also lowers
the point estimate, but the corresponding interval reaches zero. The
execution-check contrast is uncertain. Because the shared intervention-turn
protocol remains in place, the last variant does not test a workflow without
execution constraints.

\subsection{Discovery and Answer Recovery}
\label{app:discovery-answer-decomposition}

Let $D^+$ denote complete discovery and $A^+$ a correct final answer.
The underlying ledger precision, recall, and F1 are reported in
Table~\ref{tab:call-matched-discovery-crossmodel}.

\input{tables/discovery_answer_decomposition}

We examine final QA among instances with complete discovery.
Randomized views increase complete discovery for all
three systems, yet a substantial share of instances with complete discovery
still receive incorrect answers.
Conversely, some incomplete ledgers still yield the gold answer. These
counts alone do not explain why: the final agent can inspect the complete
table, including rows absent from the ledger.
The paired guidance comparison is most informative within the
complete-discovery stratum:
with the ledger held fixed, Skill produces more corrections than regressions
for every system. Complete discovery alone therefore does not determine answer recovery;
intervention remains consequential after all reviewed critical rows have
been identified.

\FloatBarrier
\subsection{Discovery Budget and Ledger Support}
\label{app:view-budget-support}

Table~\ref{tab:random5-budget-threshold} evaluates the view budget and support
threshold using five independently permuted Qwen views. For each $M$, it
averages over all
$\binom{5}{M}$ subsets rather than selecting an arbitrary prefix. A row is
retained when nominated in at least $k$ selected views. Precision, recall,
and F1 follow Appendix~\ref{app:discovery-metrics} and are macro averaged over
the 141 nonempty references and the relevant view subsets. Mean rows uses all
154 reviewed instances; Empty FP uses the 13 empty references.

\input{tables/random5_budget_threshold}

Under the union rule ($k=1$), CDR rises from 45.5\% with one view to 78.7\%
with five. The largest increment occurs between one and two views; beyond three
views, the remaining gain is 5.2 points. Recall rises from 71.8\% to 85.6\%,
while precision remains between 71.8\% and 72.7\%. The mean number of nominated
rows rises from 6.92 to 10.03, and the empty-set false-positive rate rises from
64.6\% to 76.9\%. Thus, the added views recover more required rows but also
expand the candidate set.

Requiring repeated support produces a different operating point. At $M=5$,
raising $k$ from one to two removes about two rows per ledger but reduces CDR
by 9.9 points. The same change raises precision from 72.5\% to 73.6\% but
reduces recall from 85.6\% to 80.6\%. Higher thresholds continue to shrink the
ledger and reduce recall, without consistently improving macro precision.
The end-to-end condition retains the union rule because downstream
intervention verifies each nominated row. The analysis also shows diminishing
returns by five views; it does not extrapolate to larger budgets.

\subsection{Final QA by View Count}
\label{app:view-count-qa}

We evaluate Qwen Plus final QA with the same Skill at
$M\in\{0,1,3,5\}$ (Table~\ref{tab:view-budget-qa}). Each nonzero budget uses
the first $M$ of these five permuted views; $M=0$ has no discovery ledger. QA is
evaluated on all 313 instances at each budget. Unlike the all-subset averages
in Table~\ref{tab:random5-budget-threshold}, these fixed prefixes need not
have the same discovery metrics at a given $M$. Panel (a) reports the metrics
of these exact prefixes; panel (b) reports their final QA outcomes.

\input{tables/view_budget_qa}

QA accuracy rises from 57.5\% at $M=0$ to 64.9\% at $M=5$, yielding a paired gain of
7.3 points (95\% CI $[3.2,11.9]$). The intermediate point estimates are
59.7\% at $M=1$ and 61.0\% at $M=3$; intervals for the individual increments
include zero.

%% file: tables/call_matched_discovery.tex
\begin{table}[H]
\centering\small
\setlength{\tabcolsep}{3.5pt}\renewcommand{\arraystretch}{1.06}
\caption{Discovery under an equal five-call allowance. Repeated-5 uses the
released row order; Random-5 permutes all rows independently in each call.
CDR and macro precision, recall, and F1 use 141 nonempty references.
Mean rows uses all 154 reviewed instances; Empty FP uses 13 empty references.
All rate columns are percentages.}
\label{tab:call-matched-discovery-crossmodel}
\begin{tabular}{@{}llrrrrrr@{}}\toprule
System & Views & CDR & Prec. & Rec. & F1 & Mean rows & Empty FP \\
\midrule
\multirow{2}{*}{Qwen Plus} & Repeated-5 & 58.9 & 71.0 & 77.1 & 68.9 & 9.11 & 61.5 \\
 & Random-5 & 78.7 & 72.5 & 85.6 & 74.6 & 10.03 & 76.9 \\
\addlinespace
\multirow{2}{*}{GLM-4.5-Air} & Repeated-5 & 61.0 & 60.5 & 74.6 & 62.5 & 9.11 & 84.6 \\
 & Random-5 & 66.7 & 59.7 & 76.7 & 62.3 & 10.24 & 76.9 \\
\addlinespace
\multirow{2}{*}{MiniMax-M2.5} & Repeated-5 & 38.3 & 47.1 & 48.9 & 45.4 & 4.98 & 38.5 \\
 & Random-5 & 58.2 & 65.1 & 69.4 & 64.3 & 7.14 & 23.1 \\
\bottomrule
\end{tabular}
\end{table}

%% file: tables/core_method_ablations.tex
\begin{table}[H]
    \centering
    \footnotesize
    \setlength{\tabcolsep}{4.5pt}
    \renewcommand{\arraystretch}{1.02}
    \caption{Controlled tests of the two findings with Qwen Plus. (a) QA accuracy
    on all 313 instances when intervention guidance is varied while the
    discovery ledger is held fixed within each view family. (b) Random-5 varies
    row order across five inspections, whereas Repeated-5 repeats the released
    order at the same call allowance; the QA comparison holds Skill guidance
    fixed. Differences and intervals are reported in percentage points.}
    \label{tab:core-method-ablations}
    \label{tab:call-matched-discovery}
    \label{tab:full-variant-controls}
    \begin{tabular}{@{}lrrrr@{}}
        \toprule
        \multicolumn{5}{c}{\textbf{(a) View family $\times$ guidance}} \\
        \midrule
        Views & Generic & Skill & $\Delta_{\mathrm{Skill}}$ & 95\% CI \\
        \midrule
        Repeated-5 & 172 (55.0\%) & 190 (60.7\%) & $+5.8$ & $[-0.7,12.0]$ \\
        Random-5   & 183 (58.5\%) & 203 (64.9\%) & $+6.4$ & $[1.6,11.2]$ \\
        \addlinespace[0.5em]
        \multicolumn{5}{c}{\textbf{(b) Random-5 vs. Repeated-5}} \\
        \midrule
        Metric & Repeated-5 & Random-5 & $\Delta_{\mathrm{view}}$ & 95\% CI \\
        \midrule
        Complete discovery rate & 58.9\% & 78.7\% & $+19.9$ & $[11.1,29.2]$ \\
        QA accuracy & 60.7\% & 64.9\% & $+4.2$ & $[-0.3,9.0]$ \\
        \bottomrule
    \end{tabular}
\end{table}

%% file: tables/core_ablation_artifacts.tex
\begin{table}[H]\centering\small
\setlength{\tabcolsep}{6pt}\renewcommand{\arraystretch}{1.10}
\caption{Counts of correct QA answers for the five error types and clean inputs
in the Qwen Repeated-5/Random-5 guidance factorial.}
\label{tab:core-ablation-artifacts}
\begin{tabular*}{\linewidth}{@{\extracolsep{\fill}}lrrrrr@{}}\toprule
 &  & \multicolumn{2}{c}{Generic} & \multicolumn{2}{c}{Skill} \\
\cmidrule(lr){3-4}\cmidrule(l){5-6}
Input type & $n$ & Repeated-5 & Random-5 & Repeated-5 & Random-5 \\
\midrule
Bad values & 53 & 26 & 28 & 26 & 25 \\
Clean & 53 & 46 & 49 & 47 & 49 \\
Logic & 53 & 18 & 23 & 17 & 28 \\
Formatting & 53 & 29 & 28 & 40 & 38 \\
Missingness & 53 & 25 & 25 & 31 & 30 \\
Outliers & 48 & 28 & 30 & 29 & 33 \\
\midrule
All & 313 & 172 & 183 & 190 & 203 \\
\bottomrule\end{tabular*}\end{table}

%% file: tables/skill_prompt_ablations.tex
\begin{table}[H]
\centering\small
\setlength{\tabcolsep}{5pt}
\renewcommand{\arraystretch}{1.08}
\caption{Qwen Plus final-answer accuracy on all 313 RADAR-T instances when
one part of the Intervention Skill is removed. Differences are variant
minus Full Skill, in percentage points; intervals use paired source-table
cluster resampling.}
\label{tab:skill-prompt-ablations}
\begin{tabular*}{\linewidth}{@{\extracolsep{\fill}}lrrrc@{}}
\toprule
Guidance & Correct & QA (\%) & $\Delta$ (pp) & 95\% CI (pp) \\
\midrule
Full Skill & 203/313 & 64.9 & -- & -- \\
Without recovery rules & 186/313 & 59.4 & $-5.4$ & $[-11.3,0.0]$ \\
Without formatting rules & 190/313 & 60.7 & $-4.2$ & $[-7.6,-0.4]$ \\
Without execution checks & 197/313 & 62.9 & $-1.9$ & $[-6.7,3.0]$ \\
\bottomrule
\end{tabular*}
\end{table}

%% file: tables/discovery_answer_decomposition.tex
\begin{table}[H]
    \centering
    \small
    \setlength{\tabcolsep}{3.2pt}
    \caption{Discovery completeness and final-answer outcomes on the
    141 reviewed instances with at least one critical row. Complete means that
    the union of five inspections contains every critical row. Counts
    are descriptive paired outcomes; the 13 empty-critical-row instances are
    excluded because completeness is vacuous for them.}
    \label{tab:discovery-answer-decomposition}
    \begin{tabular}{@{}lllrrrr@{}}
        \toprule
        System & Views & Guidance & $D^+A^+$ & $D^+A^-$ & $D^-A^+$ & $D^-A^-$ \\
        \midrule
        Qwen Plus & Repeated & Generic & 37 & 46 & 27 & 31 \\
         & Repeated & Skill & 51 & 32 & 28 & 30 \\
         & Random & Generic & 57 & 54 & 16 & 14 \\
         & Random & Skill & 72 & 39 & 16 & 14 \\
        \addlinespace[2pt]
        GLM-4.5-Air & Repeated & Generic & 32 & 54 & 7 & 48 \\
         & Repeated & Skill & 38 & 48 & 16 & 39 \\
         & Random & Generic & 37 & 57 & 10 & 37 \\
         & Random & Skill & 40 & 54 & 13 & 34 \\
        \addlinespace[2pt]
        MiniMax-M2.5 & Repeated & Generic & 19 & 35 & 29 & 58 \\
         & Repeated & Skill & 24 & 30 & 20 & 67 \\
         & Random & Generic & 30 & 52 & 11 & 48 \\
         & Random & Skill & 40 & 42 & 12 & 47 \\
        \bottomrule
    \end{tabular}
\end{table}

%% file: tables/random5_budget_threshold.tex
\begin{table}[H]
\centering\small
\setlength{\tabcolsep}{4pt}\renewcommand{\arraystretch}{1.04}
\caption{Qwen Plus sensitivity to the Random-view budget $M$ and ledger
support $k$. Panel (a) averages over all subsets of size $M$ from five views.
CDR and macro precision, recall, and F1 use 141 nonempty references; Mean rows
uses all 154 reviewed instances; Empty FP uses 13 empty references. Rate
columns are percentages. The deployed setting is bold.}
\label{tab:random5-budget-threshold}
\begin{tabular}{@{}crrrrrr@{}}\toprule
\multicolumn{7}{c}{\textbf{(a) View budget with union ledger ($k=1$)}} \\
$M$ & CDR & Prec. & Rec. & F1 & Mean rows & Empty FP \\
\midrule
1 & 45.5 & 71.8 & 71.8 & 68.6 & 6.92 & 64.6 \\
2 & 65.9 & 72.6 & 80.5 & 72.9 & 8.41 & 70.0 \\
3 & 73.5 & 72.7 & 83.2 & 73.9 & 9.11 & 73.1 \\
4 & 76.7 & 72.7 & 84.6 & 74.4 & 9.61 & 75.4 \\
\textbf{5} & \textbf{78.7} & \textbf{72.5} & \textbf{85.6} & \textbf{74.6} & \textbf{10.03} & \textbf{76.9} \\
\bottomrule
\end{tabular}

\vspace{0.45em}
\begin{tabular}{@{}crrrrrr@{}}\toprule
\multicolumn{7}{c}{\textbf{(b) Support threshold with five views ($M=5$)}} \\
$k$ & CDR & Prec. & Rec. & F1 & Mean rows & Empty FP \\
\midrule
\textbf{1} & \textbf{78.7} & \textbf{72.5} & \textbf{85.6} & \textbf{74.6} & \textbf{10.03} & \textbf{76.9} \\
2 & 68.8 & 73.6 & 80.6 & 73.4 & 7.97 & 69.2 \\
3 & 61.0 & 71.9 & 76.2 & 70.9 & 7.09 & 61.5 \\
4 & 44.0 & 72.4 & 68.7 & 67.0 & 5.88 & 53.8 \\
5 & 22.0 & 67.5 & 48.0 & 52.8 & 3.66 & 53.8 \\
\bottomrule
\end{tabular}
\end{table}

%% file: tables/view_budget_qa.tex
\begin{table}[H]
\centering\small
\setlength{\tabcolsep}{5pt}\renewcommand{\arraystretch}{1.06}
\caption{Qwen Plus discovery and final QA using the first $M$ of five permuted
views; $M=0$ supplies no ledger. In panel (a), CDR and macro precision,
recall, and F1 use 141 nonempty references, Mean rows uses all 154 reviewed
instances, and Empty FP uses 13 empty references. Rate columns are
percentages. Panel (b) evaluates all 313 instances; QA gains are relative to
the preceding row, with paired source-table cluster 95\% intervals.}
\label{tab:view-budget-qa}
\begin{tabular}{@{}crrrrrr@{}}\toprule
\multicolumn{7}{c}{\textbf{(a) Discovery with fixed prefixes}} \\
$M$ & CDR & Prec. & Rec. & F1 & Mean rows & Empty FP \\
\midrule
0 & 0.0 & 0.0 & 0.0 & 0.0 & 0.00 & 0.0 \\
1 & 44.7 & 72.9 & 71.1 & 68.6 & 6.51 & 61.5 \\
3 & 73.0 & 72.1 & 82.6 & 73.1 & 9.05 & 76.9 \\
5 & 78.7 & 72.5 & 85.6 & 74.6 & 10.03 & 76.9 \\
\bottomrule
\end{tabular}

\vspace{0.45em}
\begin{tabular}{@{}crrrc@{}}\toprule
\multicolumn{5}{c}{\textbf{(b) Final QA}} \\
$M$ & Correct & QA (\%) & Gain (pp) & 95\% CI \\
\midrule
0 & 180/313 & 57.5 & -- & -- \\
1 & 187/313 & 59.7 & $+2.2$ & $[-3.0,7.9]$ \\
3 & 191/313 & 61.0 & $+1.3$ & $[-4.0,6.3]$ \\
5 & 203/313 & 64.9 & $+3.8$ & $[-0.9,8.3]$ \\
\bottomrule
\end{tabular}
\end{table}

%% file: appendices/appendix_g_uncertainty.tex
\section{Uncertainty Estimates}
\label{app:uncertainty}

We report paired confidence intervals for the geometry study
(Section~\ref{sec:geometry}), the table-state comparison
(Section~\ref{sec:intervention-motivation}), and GBDI
(Section~\ref{sec:end-to-end-validation}). The resampling unit follows the
shared source of dependence within each evaluation.

\subsection{Cluster Bootstrap}
\label{app:cluster-bootstrap}

We resample clusters to retain dependence among related observations
\citep{field2007bootstrap}. The reviewed geometry analysis uses 20,000 base-task cluster-bootstrap
samples over 29 base tasks. All instances and backgrounds associated with a
sampled task are retained together. Complete discovery rate (CDR) contrasts
use an instance-weighted estimand.

The 154-instance table-state comparison uses 10,000 source-table cluster
bootstrap samples over 24 source tables. End-to-end QA comparisons use
10,000 source-table cluster bootstrap samples with the task-to-source mapping
from RADAR~\citep{gu2025radar}. None of these procedures treats repeated views as
independent instances.

For a contrast between conditions $A$ and $B$, let $a_i$ denote the paired
difference in binary outcomes and $b_i=1$. For QA, the outcome is answer
correctness; for discovery, it is the indicator in
Equation~\ref{eq:complete-discovery-rate}. For geometry, paired contrasts are
first averaged within each instance.
With $m_s^{(b)}$ denoting the number of times cluster $s$ appears in bootstrap
sample $b$, the estimate is
\begin{equation}
 \widehat\Delta^{(b)} = 100\,
 \frac{\sum_s m_s^{(b)}\sum_{i\in s}a_i}
      {\sum_s m_s^{(b)}\sum_{i\in s}b_i}.
\end{equation}
The reported endpoints are the 2.5th and 97.5th percentiles of these estimates.
Clustering changes the resampling unit, not the instance or reference-row
weighting of the reported estimand. Intervals are pointwise rather than simultaneous across
all systems and contrasts.
For position, the q5--q1 contrast pairs the two endpoint layouts within each
instance and background, using all 60 instances. We also report a descriptive
linear trend: for each instance and background, we regress the binary
complete-discovery outcome on the ordinal band index (1 through 5).
Endpoint differences and slopes are averaged over the three backgrounds
within each instance before cluster resampling. For dispersion, the
Wide--Compact contrast uses 59 instances and averages paired differences over
the three mean positions and three backgrounds within each instance.

\input{tables/geometry_uncertainty}

Invalid submission counts for position (out of 900 per mode) are 0/0 for Qwen,
8/207 for MiniMax, and 0/0 for GPT, with Direct listed first and Code Agent
second. For dispersion (out of 1,593 per mode), the corresponding counts are
1/0, 11/386, and 2/1, respectively. Invalid outputs are scored as empty
predictions and remain in the denominators; all planned trajectories are present.

\subsection{Direct vs. Code Agent}
\label{app:interface-attenuation}

For each instance, value of $\mu$, and background, we pair Direct and Code
Agent outputs on the same layout and compute
$\operatorname{CDR}(\text{Wide})-\operatorname{CDR}(\text{Compact})$. The Code
Agent effect minus the Direct effect is averaged over backgrounds and values of $\mu$
within each instance before base-task resampling. This paired contrast
compares the effects across modes without relying on whether the separate
within-mode intervals include zero. A positive value indicates less
Compact-to-Wide degradation under the Code Agent. Means are computed over $\mu$ values
and backgrounds within each of the 59 instances, preserving paired outcomes
from both modes.

\input{tables/geometry_interface_contrasts}

The mean CDR contrast favors the Code Agent for all three systems. The paired
intervals exclude zero for Qwen and GPT; the MiniMax interval includes zero.
Position-specific differences are positive for Qwen and GPT; MiniMax varies
across values of $\mu$. These comparisons retain invalid
outputs in their original denominators and compare the complete modes,
which differ in both programmatic access and response allowance.

\subsection{Table-State Contrasts}
\label{app:table-state-contrasts}

\input{tables/intervention_information_uncertainty}

Within each mode, the three states use the same instances, questions, system
prompt, response allowance, and answer scorer. The
Discovery--Raw intervals include or touch zero for all three systems in both
modes. Action--Discovery intervals exclude zero for MiniMax and GPT under
Direct and for all three systems under the Code Agent, but not for Qwen under Direct. Because
Action provides an already repaired table, this contrast does not isolate
the effect of an action label.

\subsection{GBDI Discovery and QA}
\label{app:gbdi-discovery-qa}

CDR contrasts use 141 reviewed instances and resample 34 base tasks; QA
contrasts resample 27 source tables. Random-5 uses five independently permuted
views. Repeated-5 performs five independent inspections of the released row
order, with the inspection and final QA budgets held fixed. GPT has no
Repeated-5 arm.

\input{tables/discovery_uncertainty}
\input{tables/main_results_gbea}
\input{tables/qa_uncertainty}

The QA intervals for GBDI relative to the Code Agent baseline exclude zero for
Qwen, GLM, MiniMax, and DeepSeek-V4.1-Flash, but not GPT. QA differences between Random-5
and Repeated-5 under equal call budgets are less certain. The corresponding
CDR intervals exclude zero for Qwen and MiniMax, but not GLM.

%% file: tables/geometry_uncertainty.tex
\begin{table}[H]
\centering\small
\setlength{\tabcolsep}{3pt}\renewcommand{\arraystretch}{1.10}
\caption{Position and dispersion contrasts in complete discovery rate.
Entries give estimates with pointwise 95\% confidence intervals in brackets.
Endpoint is q5$-$q1; trend is the slope per ordinal position band.
Dispersion is Wide$-$Compact. Differences are in percentage points;
trends are in percentage points per band.}
\label{tab:geometry-uncertainty}
\begin{tabular*}{\linewidth}{@{\extracolsep{\fill}}llrrr@{}}\toprule
 & & \multicolumn{2}{c}{Position (60 instances)} & Dispersion (59 instances) \\
\cmidrule(lr){3-4}\cmidrule(l){5-5}
System & Mode & Endpoint & Trend & Wide$-$Compact \\
\midrule
Qwen Plus & Direct & $+11.7\;[+4.0, +20.2]$ & $+2.5\;[+0.7, +4.5]$ & $-16.2\;[-22.4, -10.5]$ \\
 & Code Agent & $+7.2\;[-0.6, +15.8]$ & $+1.3\;[-0.4, +3.1]$ & $-2.3\;[-6.6, +1.8]$ \\
\addlinespace[0.4em]
MiniMax-M2.5 & Direct & $+11.7\;[+4.6, +19.1]$ & $+1.9\;[+0.3, +3.7]$ & $-9.6\;[-13.0, -6.2]$ \\
 & Code Agent & $+5.0\;[-2.6, +13.6]$ & $+1.5\;[-0.1, +3.4]$ & $-3.8\;[-9.3, +1.3]$ \\
\addlinespace[0.4em]
GPT-5 Mini & Direct & $+16.7\;[+8.1, +26.9]$ & $+4.1\;[+2.0, +6.3]$ & $-7.9\;[-13.0, -2.5]$ \\
 & Code Agent & $+3.3\;[-3.5, +10.3]$ & $+0.5\;[-0.6, +1.7]$ & $-2.1\;[-5.0, +1.3]$ \\
\bottomrule\end{tabular*}
\end{table}

%% file: tables/geometry_interface_contrasts.tex
\begin{table}[H]
\centering\footnotesize
\setlength{\tabcolsep}{3.2pt}\renewcommand{\arraystretch}{1.10}
\caption{Code Agent vs. Direct differences in the Wide vs. Compact CDR effect,
in percentage points.}\label{tab:geometry-interface-contrasts}
\begin{tabular}{@{}lrrrrr@{}}\toprule
System & Mean & 95\% CI & $\mu=25\%$ & $\mu=50\%$ & $\mu=75\%$ \\
\midrule
Qwen Plus & +13.9 & [+8.4, +20.0] & +19.2 & +16.9 & +5.6 \\
MiniMax-M2.5 & +5.8 & [-0.2, +11.7] & +4.5 & +13.0 & +0.0 \\
GPT-5 Mini & +5.8 & [+0.4, +11.5] & +3.4 & +11.9 & +2.3 \\
\bottomrule\end{tabular}
\end{table}

%% file: tables/intervention_information_uncertainty.tex
\begin{table}[H]
\centering\footnotesize
\setlength{\tabcolsep}{3.0pt}\renewcommand{\arraystretch}{1.08}
\caption{Final-answer accuracy (\%) and paired table-state contrasts on the
154 reviewed instances. Differences and source-table cluster 95\% intervals
are reported in percentage points. Raw uses the original table, Discovery uses the
error-marked table, and Action uses the repaired table.}
\label{tab:intervention-information-uncertainty}
\begin{tabular*}{\linewidth}{@{\extracolsep{\fill}}llrrrcc@{}}
\toprule
System & Mode & Raw & Discovery & Action & Discovery vs. Raw & Action vs. Discovery \\
\midrule
\multirow{2}{*}{Qwen Plus}
 & Direct & 16.2 & 18.8 & 29.2 & $+2.6\;[-5.5,10.1]$ & $+10.4\;[-1.2,23.3]$ \\
 & Code Agent & 46.8 & 56.5 & 98.1 & $+9.7\;[0.0,19.5]$ & $+41.6\;[34.1,49.2]$ \\
\addlinespace
\multirow{2}{*}{MiniMax-M2.5}
 & Direct & 35.7 & 31.8 & 67.5 & $-3.9\;[-16.4,9.0]$ & $+35.7\;[22.4,47.4]$ \\
 & Code Agent & 25.3 & 33.8 & 92.9 & $+8.4\;[-0.6,16.8]$ & $+59.1\;[49.5,68.9]$ \\
\addlinespace
\multirow{2}{*}{GPT-5 Mini}
 & Direct & 18.8 & 22.7 & 37.0 & $+3.9\;[-3.3,11.5]$ & $+14.3\;[5.7,24.1]$ \\
 & Code Agent & 47.4 & 58.4 & 97.4 & $+11.0\;[-2.4,24.1]$ & $+39.0\;[29.0,49.3]$ \\
\bottomrule
\end{tabular*}
\end{table}

%% file: tables/discovery_uncertainty.tex
\begin{table}[H]
\centering\small
\setlength{\tabcolsep}{8pt}\renewcommand{\arraystretch}{1.12}
\caption{Random-5 vs. Repeated-5 CDR on 141 reviewed instances. Differences
are in percentage points.}
\label{tab:discovery-uncertainty}
\begin{tabular}{@{}lrr@{}}\toprule
System & $\Delta$ CDR & 95\% CI \\
\midrule
Qwen Plus & $+19.86$ & $[+11.11,\,+29.23]$ \\
GLM-4.5-Air & $+5.67$ & $[-1.44,\,+13.48]$ \\
MiniMax-M2.5 & $+19.86$ & $[+9.59,\,+29.73]$ \\
\bottomrule
\end{tabular}
\end{table}

%% file: tables/main_results_gbea.tex
\begin{table}[!htbp]
    \centering\small
    \setlength{\tabcolsep}{4pt}
    \renewcommand{\arraystretch}{1.10}
    \caption{End-to-end QA accuracy (\%) on all 313 RADAR-T instances.
    GBDI combines five randomized inspections with intervention guidance
    before final Code Agent QA. $\Delta_{\mathrm{Direct}}$ and $\Delta_{\mathrm{Code}}$
    denote the gains over each baseline; the paired 95\% interval is for
    $\Delta_{\mathrm{Code}}$. All differences are reported in percentage points.}
    \label{tab:main-gbea-results}
    \begin{tabular*}{\linewidth}{@{\extracolsep{\fill}}lrrrrrl@{}}
        \toprule
        & \multicolumn{2}{c}{Baseline} & \multicolumn{1}{c}{Test condition} & \multicolumn{3}{c}{Gain (pp)} \\
        \cmidrule(lr){2-3}\cmidrule(lr){4-4}\cmidrule(l){5-7}
        System & Direct & Code Agent & GBDI & $\Delta_{\mathrm{Direct}}$ & $\Delta_{\mathrm{Code}}$ & 95\% CI \\
        \midrule
        Qwen Plus & 11.8 & 54.0 & 64.9 & +53.0 & +10.9 & $[5.3,\,16.5]$ \\
        GLM-4.5-Air & 4.5 & 41.2 & 49.5 & +45.0 & +8.3 & $[2.9,\,14.0]$ \\
        MiniMax-M2.5 & 44.4 & 32.6 & 51.1 & +6.7 & +18.5 & $[11.9,\,25.6]$ \\
        GPT-5 Mini & 27.2 & 56.9 & 60.7 & +33.5 & +3.8 & $[-2.5,\,9.8]$ \\
        DeepSeek-V4.1-Flash & 49.8 & 59.7 & 67.7 & +17.9 & +8.0 & $[2.9,\,12.5]$ \\
        \bottomrule
    \end{tabular*}
\end{table}

%% file: tables/qa_uncertainty.tex
\begin{table}[H]
\centering\small
\setlength{\tabcolsep}{8pt}\renewcommand{\arraystretch}{1.12}
\caption{Paired GBDI QA gains on 313 instances (percentage points).}\label{tab:qa-uncertainty}
\begin{tabular}{@{}llrr@{}}\toprule
System & Reference & $\Delta$ & 95\% CI \\
\midrule
Qwen Plus & Code Agent & $+10.86$ & $[+5.26,\,+16.51]$ \\
 & Repeated + Skill & $+4.15$ & $[-0.32,\,+9.03]$ \\
\addlinespace
MiniMax-M2.5 & Code Agent & $+18.53$ & $[+11.87,\,+25.62]$ \\
 & Repeated + Skill & $+4.79$ & $[+0.00,\,+9.59]$ \\
\addlinespace
GPT-5 Mini & Code Agent & $+3.83$ & $[-2.50,\,+9.80]$ \\
\addlinespace
DeepSeek-V4.1-Flash & Code Agent & $+7.99$ & $[+2.94,\,+12.54]$ \\
\addlinespace
GLM-4.5-Air & Code Agent & $+8.31$ & $[+2.92,\,+13.97]$ \\
 & Repeated + Skill & $+1.28$ & $[-3.61,\,+6.17]$ \\
\bottomrule
\end{tabular}
\end{table}

%% file: appendices/appendix_e_prompts.tex
\clearpage
\section{Prompts and Output Schemas}
\label{app:prompts}
\raggedbottom

The prompt components below follow the geometry diagnostic
(Section~\ref{sec:geometry}), the table-state comparison
(Section~\ref{sec:intervention-motivation}), and GBDI
(Section~\ref{sec:end-to-end-validation}). The boxes reproduce the text used
in the reported evaluations. Role labels distinguish system instructions
from user inputs; bracketed uppercase fields denote
instance-specific content. We omit redundant internal titles, format headings,
lists, and command examples, and normalize minor encoding and typographical
artifacts for readability.

\subsection{Geometry Discovery}
\label{app:geometry-prompts}

Direct discovery requests a JSON object with \texttt{candidates} and
\texttt{unresolved}. Each candidate has \texttt{row\_id}, \texttt{columns}, and
\texttt{evidence}; there are no separate observed-value or relevance fields.
The Code Agent uses the same discovery objective with the command interface
shown below.

  \ifcsname promptkeep@geometry_direct_system\endcsname
    \tcbset{promptpagination/.style={unbreakable}}%
  \else
    \tcbset{promptpagination/.style={}}%
  \fi
  \begin{tcolorbox}[paperprompt,
    promptpagination,
    title={\promptcardtitle{system}{Direct discovery}},
    title after break={\promptcardtitle{system}{Direct discovery}\hfill\normalfont\small continued}]
  \input{prompts/typeset/geometry_direct_system.tex}
  \end{tcolorbox}
  \ifcsname promptkeep@geometry_direct_user\endcsname
    \tcbset{promptpagination/.style={unbreakable}}%
  \else
    \tcbset{promptpagination/.style={}}%
  \fi
  \begin{tcolorbox}[paperprompt,
    promptpagination,
    title={\promptcardtitle{user}{Direct discovery input}},
    title after break={\promptcardtitle{user}{Direct discovery input}\hfill\normalfont\small continued}]
  \input{prompts/typeset/geometry_direct_user.tex}
  \end{tcolorbox}

Code Agent discovery uses two YAML commands. The \texttt{python} command
supports inspection, and \path{submit_localization} submits the JSON payload.
The runtime allows at most five response rounds, maintains a persistent
namespace, and restores the visible \texttt{df} after each inspection command.
This discovery endpoint differs from the \texttt{done} submission used by the
QA agent.

  \ifcsname promptkeep@geometry_code_system\endcsname
    \tcbset{promptpagination/.style={unbreakable}}%
  \else
    \tcbset{promptpagination/.style={}}%
  \fi
  \begin{tcolorbox}[paperprompt,
    promptpagination,
    title={\promptcardtitle{system}{Code Agent discovery}},
    title after break={\promptcardtitle{system}{Code Agent discovery}\hfill\normalfont\small continued}]
  \input{prompts/typeset/geometry_code_system.tex}
  \end{tcolorbox}
  \ifcsname promptkeep@geometry_code_user\endcsname
    \tcbset{promptpagination/.style={unbreakable}}%
  \else
    \tcbset{promptpagination/.style={}}%
  \fi
  \begin{tcolorbox}[paperprompt,
    promptpagination,
    title={\promptcardtitle{user}{Code Agent discovery input}},
    title after break={\promptcardtitle{user}{Code Agent discovery input}\hfill\normalfont\small continued}]
  \input{prompts/typeset/geometry_code_user.tex}
  \end{tcolorbox}

\subsection{Table-State Diagnostic}
\label{app:table-state-prompts}

Raw uses the standard table--question prompt for each mode. Discovery and
Action retain the mode-specific system prompt and prepend the following
notices to the user turn. Discovery marks erroneous table values as
\texttt{<<ERROR:value>>}; Action supplies the table after reviewed edits and
drops have been applied. The notices are identical across Direct and Code
Agent.

  \ifcsname promptkeep@diagnostic_marked_notice\endcsname
    \tcbset{promptpagination/.style={unbreakable}}%
  \else
    \tcbset{promptpagination/.style={}}%
  \fi
  \begin{tcolorbox}[paperprompt,
    promptpagination,
    title={\promptcardtitle{user}{Discovery condition}},
    title after break={\promptcardtitle{user}{Discovery condition}\hfill\normalfont\small continued}]
  \input{prompts/typeset/diagnostic_marked_notice.tex}
  \end{tcolorbox}
  \ifcsname promptkeep@diagnostic_intervened_notice\endcsname
    \tcbset{promptpagination/.style={unbreakable}}%
  \else
    \tcbset{promptpagination/.style={}}%
  \fi
  \begin{tcolorbox}[paperprompt,
    promptpagination,
    title={\promptcardtitle{user}{Action condition}},
    title after break={\promptcardtitle{user}{Action condition}\hfill\normalfont\small continued}]
  \input{prompts/typeset/diagnostic_intervened_notice.tex}
  \end{tcolorbox}

\subsection{Multi-View Inspection Schemas}
\label{app:inspection-schemas}

Each view is processed in a fresh context under Random-$M$ or Repeated-$M$,
as specified in Appendix~\ref{app:gbdi-protocol}. Within a system, the two
conditions share the inspection prompt and parser. The evaluation uses
the following two schemas.

\paragraph{Qwen Plus, GLM, MiniMax, and DeepSeek-V4.1-Flash.}
The output contains \texttt{row\_id}, a \texttt{columns} list, and
\texttt{evidence}, together with the top-level \texttt{unresolved} list.

  \ifcsname promptkeep@discovery_three_models_system\endcsname
    \tcbset{promptpagination/.style={unbreakable}}%
  \else
    \tcbset{promptpagination/.style={}}%
  \fi
  \begin{tcolorbox}[paperprompt,
    promptpagination,
    title={\promptcardtitle{system}{Multi-view discovery}},
    title after break={\promptcardtitle{system}{Multi-view discovery}\hfill\normalfont\small continued}]
  \input{prompts/typeset/discovery_three_models_system.tex}
  \end{tcolorbox}
  \ifcsname promptkeep@discovery_three_models_user\endcsname
    \tcbset{promptpagination/.style={unbreakable}}%
  \else
    \tcbset{promptpagination/.style={}}%
  \fi
  \begin{tcolorbox}[paperprompt,
    promptpagination,
    title={\promptcardtitle{user}{Multi-view discovery input}},
    title after break={\promptcardtitle{user}{Multi-view discovery input}\hfill\normalfont\small continued}]
  \input{prompts/typeset/discovery_three_models_user.tex}
  \end{tcolorbox}

\paragraph{GPT-5 Mini.}
The prompt requests observed values and query relevance alongside the candidate
rows, columns, and evidence. The parser accepts blank
observed values for missing cells and requires
nonempty relevance and evidence text. It normalizes the singular
\texttt{column} field to a column list before aggregation.

  \ifcsname promptkeep@discovery_gpt\endcsname
    \tcbset{promptpagination/.style={unbreakable}}%
  \else
    \tcbset{promptpagination/.style={}}%
  \fi
  \begin{tcolorbox}[paperprompt,
    promptpagination,
    title={\promptcardtitle{system}{GPT-5 Mini discovery}},
    title after break={\promptcardtitle{system}{GPT-5 Mini discovery}\hfill\normalfont\small continued}]
  \input{prompts/typeset/discovery_gpt.tex}
  \end{tcolorbox}
  \ifcsname promptkeep@discovery_gpt_user\endcsname
    \tcbset{promptpagination/.style={unbreakable}}%
  \else
    \tcbset{promptpagination/.style={}}%
  \fi
  \begin{tcolorbox}[paperprompt,
    promptpagination,
    title={\promptcardtitle{user}{GPT-5 Mini discovery input}},
    title after break={\promptcardtitle{user}{GPT-5 Mini discovery input}\hfill\normalfont\small continued}]
  \input{prompts/typeset/discovery_gpt_user.tex}
  \end{tcolorbox}

\subsection{Evidence Ledgers}
\label{app:evidence-processing}

For a fixed instance and row $r$, let $c_M(r)=\sum_{t=1}^{M}\mathbf{1}[r\in\widehat{\mathcal R}_t]$
count the views that nominate the row. A ledger entry has the conceptual form
\begin{equation}
L_M(r)=\bigl[r,\;\mathcal{J}_r,\;\mathcal{X}_r,\;c_M(r),\;\mathcal{E}_r\bigr],
\label{eq:evidence-ledger}
\end{equation}
where $\mathcal{J}_r$ contains the named columns, $\mathcal{X}_r$ their visible
values, and $\mathcal{E}_r$ the retained observations and any supplied relevance
claims. Entries are included whenever $c_M(r)\ge1$. The values in
$\mathcal{X}_r$ are copied from the canonical table; statements in
$\mathcal{E}_r$ remain system-generated claims that the final agent must verify.

Both implementations retain the union of valid row nominations and count
each row at most once per view. Ledger construction uses only system
nominations and requires no support beyond one valid nomination.
The text-processing differences described below are held constant across view
families within each system.

\paragraph{Qwen, GLM, MiniMax, and DeepSeek-V4.1-Flash.}
Before constructing the final agent's input, the program unions valid column
names, normalizes whitespace, and removes duplicate observations in view order.
It also removes observations that match predefined, case-insensitive expressions
suggesting repair actions or replacement values, such as ``edit,'' ``drop,''
``should be,'' or ``correct value.'' This filtering removes the matching
observation, while retaining the row nomination, named columns, and support count
for verification. The \texttt{unresolved} field is not forwarded, and rows are
ordered by decreasing support.

The resulting ledger is passed to the final agent in the user-message block
below. Each entry identifies the nominated row and includes its named columns,
per-view vote mask, visible cell values, and retained observations. A vote mask uses
\texttt{1} for nomination, \texttt{0} for absence, and \texttt{?} for an
unparseable view. Empty candidate sets produce no appended ledger.

  \ifcsname promptkeep@ledger_three_models\endcsname
    \tcbset{promptpagination/.style={unbreakable}}%
  \else
    \tcbset{promptpagination/.style={}}%
  \fi
  \begin{tcolorbox}[paperprompt,
    promptpagination,
    title={\promptcardtitle{user}{Discovery ledger}},
    title after break={\promptcardtitle{user}{Discovery ledger}\hfill\normalfont\small continued}]
  \input{prompts/typeset/ledger_three_models.tex}
  \end{tcolorbox}

\paragraph{GPT-5 Mini.}
GPT merges valid nominations for the same row, unions columns, and deduplicates
relevance and evidence text. The resulting ledger includes visible cell values
and retains all evidence text. Rows are ordered by decreasing support.
The ledger presents nominations as claims to verify and
reports when no row was nominated. Protocol matching therefore applies within
each system; ledger format differs between GPT-5 Mini and the other four
systems.

\subsection{Intervention Guidance}
\label{app:intervention-guidance}

Both Random-5 and Repeated-5 use the Intervention Skill below. The guidance
ablation replaces it with the generic guidance box while retaining the same
intervention turn and final QA budget.

  \ifcsname promptkeep@generic_guidance\endcsname
    \tcbset{promptpagination/.style={unbreakable}}%
  \else
    \tcbset{promptpagination/.style={}}%
  \fi
  \begin{tcolorbox}[paperprompt,
    promptpagination,
    title={\promptcardtitle{system}{Generic intervention guidance}},
    title after break={\promptcardtitle{system}{Generic intervention guidance}\hfill\normalfont\small continued}]
  \input{prompts/typeset/generic_guidance.tex}
  \end{tcolorbox}
  \ifcsname promptkeep@intervention_skill\endcsname
    \tcbset{promptpagination/.style={unbreakable}}%
  \else
    \tcbset{promptpagination/.style={}}%
  \fi
  \begin{tcolorbox}[paperprompt,
    promptpagination,
    title={\promptcardtitle{system}{Intervention Skill}},
    title after break={\promptcardtitle{system}{Intervention Skill}\hfill\normalfont\small continued}]
  \input{prompts/typeset/intervention_skill.tex}
  \end{tcolorbox}

\Needspace{10\baselineskip}
\subsection{QA Conversation}
\label{app:qa-conversation}

The system message combines the RADAR Code Agent base prompt, the selected
guidance, and the intervention-turn contract, in that order.
The command schema and permissions remain
those of the base prompt, while the selected guidance specifies the evidence
required for allowed actions. The base prompt includes the \texttt{python} and
\texttt{done} schemas.

  \ifcsname promptkeep@qa_base_system\endcsname
    \tcbset{promptpagination/.style={unbreakable}}%
  \else
    \tcbset{promptpagination/.style={}}%
  \fi
  \begin{tcolorbox}[paperprompt,
    promptpagination,
    title={\promptcardtitle{system}{Code Agent base prompt}},
    title after break={\promptcardtitle{system}{Code Agent base prompt}\hfill\normalfont\small continued}]
  \input{prompts/typeset/qa_base_system.tex}
  \end{tcolorbox}
  \ifcsname promptkeep@intervention_preturn\endcsname
    \tcbset{promptpagination/.style={unbreakable}}%
  \else
    \tcbset{promptpagination/.style={}}%
  \fi
  \begin{tcolorbox}[paperprompt,
    promptpagination,
    title={\promptcardtitle{system}{Intervention-turn contract}},
    title after break={\promptcardtitle{system}{Intervention-turn contract}\hfill\normalfont\small continued}]
  \input{prompts/typeset/intervention_preturn.tex}
  \end{tcolorbox}
  \ifcsname promptkeep@qa_user\endcsname
    \tcbset{promptpagination/.style={unbreakable}}%
  \else
    \tcbset{promptpagination/.style={}}%
  \fi
  \begin{tcolorbox}[paperprompt,
    promptpagination,
    title={\promptcardtitle{user}{Initial QA input}},
    title after break={\promptcardtitle{user}{Initial QA input}\hfill\normalfont\small continued}]
  \input{prompts/typeset/qa_user.tex}
  \end{tcolorbox}

The blank-ledger diagnostic omits the ledger block while retaining the guidance,
contract, and canonical table. It therefore measures the effect of ledger availability within
the same candidate-guided workflow. Before the intervention turn, the
runtime initializes \texttt{work\_df = df.copy()} and an empty
\texttt{intervention\_log}. It executes the returned Python command, reports
its observation and contract status, and appends the continuation below.
The runtime would restore the initial value if either object had an invalid
type, while retaining the trajectory in the primary score. This fallback was
not triggered in any of the five reported 313-instance Random runs: both
objects had valid types after every reserved turn. Final QA continues in the
same conversation and namespace.

  \ifcsname promptkeep@qa_continuation\endcsname
    \tcbset{promptpagination/.style={unbreakable}}%
  \else
    \tcbset{promptpagination/.style={}}%
  \fi
  \begin{tcolorbox}[paperprompt,
    promptpagination,
    title={\promptcardtitle{user}{Continuation after intervention}},
    title after break={\promptcardtitle{user}{Continuation after intervention}\hfill\normalfont\small continued}]
  \input{prompts/typeset/qa_continuation.tex}
  \end{tcolorbox}

Up to five further responses may use ordinary \texttt{python} commands or submit
\texttt{done} with \texttt{kwargs.answer}. Python observations are returned
using the benchmark observation template. Benchmark QA scoring credits only a
valid final submission, not an answer reconstructed from code or stdout.
Neither the ledger nor the internal working table is an additional benchmark
output.

Our code is available at
\url{https://github.com/645-t/GBDI-ICLR-2027}.

%% file: prompts/typeset/geometry_direct_system.tex
\promptheading{SETTING}
You are a diagnostic table inspector. Inspect the visible table only for the downstream question. Do not repair the table and do not answer the question.\par
Identify only rows for which the visible table provides concrete evidence that a query-relevant value or relation is invalid or unusable. Do not list a row merely because it is unusual, irrelevant, extreme, or difficult to interpret. When evidence suggests a problem but does not identify one row, record it as unresolved instead of listing every possible row.\par
Return exactly one JSON object and no markdown:\par
\promptcodefile{prompts/typeset/geometry_direct_system_code1.txt}
An empty candidates list is valid. Gold answers, hidden references, artifact labels, and external facts are unavailable.\par

%% file: prompts/typeset/geometry_direct_user.tex
\promptfield{Downstream question:}{[QUESTION]}
\promptfield{Visible table (also available as pandas dataframe \promptinline{df} in repair mode):}{[TABLE VIEW]}
All cells are object dtype. \promptinline{\_\_row\_id\_\_} is runtime metadata, not a source-table field.\par

%% file: prompts/typeset/geometry_code_system.tex
\promptheading{SETTING}
You are a diagnostic table inspector. Inspect the visible table only for the downstream question. Do not repair the table and do not answer the question.\par
Identify only rows for which the visible table provides concrete evidence that a query-relevant value or relation is invalid or unusable. Do not list a row merely because it is unusual, irrelevant, extreme, or difficult to interpret. When evidence suggests a problem but does not identify one row, record it as unresolved instead of listing every possible row.\par
You may inspect the complete pandas dataframe \promptinline{df} with read-only Python. It contains the exact visible CSV and all cells are object dtype. Use Python when it helps check complete-column patterns, cross-column relations, masks, or candidate coverage. \promptinline{pd}, \promptinline{np}, \promptinline{json}, \promptinline{math}, \promptinline{re}, and \promptinline{statistics} are already available. Variables persist across inspection turns, while \promptinline{df} is restored after each command.\par
On every turn return exactly one YAML command and no prose.\par
To inspect the table:\par
\promptcodefile{prompts/typeset/geometry_code_system_code1.txt}
To finish:\par
\promptcodefile{prompts/typeset/geometry_code_system_code2.txt}
An empty candidates list is valid. Candidate row ids and columns must exist in the visible table. Gold answers, hidden references, artifact labels, and external facts are unavailable.\par

%% file: prompts/typeset/geometry_code_user.tex
\promptfield{Downstream question:}{[QUESTION]}
\promptfield{Visible table (also available as pandas dataframe \promptinline{df}):}{[TABLE VIEW]}
All cells are object dtype. \promptinline{\_\_row\_id\_\_} is runtime metadata, not a source-table field.\par

%% file: prompts/typeset/diagnostic_marked_notice.tex
The values marked \textless{}\textless{}ERROR:...\textgreater{}\textgreater{} have been verified as erroneous and relevant to the question; the text inside each marker is the original erroneous value, not a correction. Treat these markings as facts when answering: correct a marked value when the table supports a reliable replacement, or exclude its row if it cannot be repaired.\par

%% file: prompts/typeset/diagnostic_intervened_notice.tex
The table below has already been repaired for this question. All reviewed errors relevant to the answer have been corrected, and rows that could not be repaired have been removed. Treat the supplied table as final and answer the question without making further changes.\par

%% file: prompts/typeset/discovery_three_models_system.tex
\promptheading{SETTING}
You are a diagnostic table inspector. Inspect the visible table only for the downstream question. Do not repair the table and do not answer the question.\par
Identify only rows for which the visible table provides concrete evidence that a query-relevant value or relation is invalid or unusable. Do not list a row merely because it is unusual, irrelevant, extreme, or difficult to interpret. When evidence suggests a problem but does not identify one row, record it as unresolved instead of listing every possible row.\par
Return exactly one JSON object and no markdown:\par
\promptcodefile{prompts/typeset/discovery_three_models_system_code1.txt}
An empty candidates list is valid. Gold answers, hidden references, artifact labels, and external facts are unavailable.\par

%% file: prompts/typeset/discovery_three_models_user.tex
\promptfield{Downstream question:}{[QUESTION]}
\promptfield{Visible table:}{[TABLE VIEW]}
All cells are object dtype. \promptinline{\_\_row\_id\_\_} is runtime metadata identifying the canonical source row; it is not a source-table field.\par

%% file: prompts/typeset/discovery_gpt.tex
You are a diagnostic table inspector. You are given a downstream question and a complete table view. The \promptinline{\_\_row\_id\_\_} column contains stable, nonordinal row identifiers used to map records across views. It is not a source-table field and does not encode row order.\par
Identify only rows whose visible contents provide concrete evidence of an unreliable value or relation that can affect the requested computation. Do not nominate a row merely because it is unusual, extreme, irrelevant, excluded by an ordinary query predicate, or difficult to interpret.\par
For each supported candidate, report the row ID, one affected visible column, the displayed value in that cell, why that field can enter the requested computation, and one concise observation that can be checked in the displayed table. Emit a separate record for each affected column when needed.\par
Do not repair the table or answer the question. Do not propose a replacement or choose EDIT, DROP, or NO-OP. Gold answers, recovered tables, artifact labels, task identifiers, and predictions from other views are unavailable.\par
Return exactly one JSON object and no Markdown:\par
\promptcodefile{prompts/typeset/discovery_gpt_code1.txt}
An empty \promptinline{candidates} list is valid.\par

%% file: prompts/typeset/discovery_gpt_user.tex
\promptfield{Downstream question:}{[QUESTION]}
\promptfield{Complete visible table view:}{[TABLE VIEW]}
All source-table cells are displayed as strings. \promptinline{\_\_row\_id\_\_} is stable, nonordinal runtime metadata used only to identify rows across views.\par

%% file: prompts/typeset/ledger_three_models.tex
Five independent inspections examined semantically equivalent views of this same table. Every nominated canonical row is retained below, including single-view nominations. Consensus is an inspection-priority signal, not proof that a row is wrong and not authorization to change or exclude data.\par
Before answering:\par
\promptitem{0.6}{1.}{establish the exact query operation and answer scope;}
\promptitem{0.6}{2.}{inspect every ledger row in the complete canonical table;}
\promptitem{0.6}{3.}{use the vote pattern and observations only to prioritize verification;}
\promptitem{0.6}{4.}{ignore nominations not supported by the visible table; and}
\promptitem{0.6}{5.}{directly compute and submit the final answer.}
The ledger contains no Gold labels, prescribed operations, replacements, recovered table, Repair Skill, or answer.\par

%% file: prompts/typeset/generic_guidance.tex
\promptitem{0.6}{\textbullet}{Treat every ledger coordinate and observation as a lead to verify against the complete visible table, not as proof of corruption or a prescribed operation.}
\promptitem{0.6}{\textbullet}{Preserve the literal answer program required by the question.}
\promptitem{0.6}{\textbullet}{EDIT only when visible evidence determines one exact replacement.}
\promptitem{0.6}{\textbullet}{DROP only when a record required by the answer program is demonstrably unusable and no unique replacement is justified.}
\promptitem{0.6}{\textbullet}{Otherwise use NO-OP, including for usable, irrelevant, unusual-but-valid, or ambiguous data.}
\promptitem{0.6}{\textbullet}{Apply justified operations to \promptinline{work\_df} before rerunning the unchanged answer program.}

%% file: prompts/typeset/intervention_skill.tex
\promptheading{Entry Gate}
\promptitem{0.6}{\textbullet}{Complete the original question-answering task in one code-agent context.}
\promptitem{0.6}{\textbullet}{Treat the balanced-view ledger as a fallible inspection aid. A coordinate or observation nominates a candidate but never proves corruption or prescribes an operation.}
\promptitem{0.6}{\textbullet}{Before using an artifact rule, verify one of its direct visible preconditions: an answer-active missing marker, a hard invalid value, or a representation that blocks parsing of an otherwise recoverable value.}
\promptitem{0.6}{\textbullet}{If none of those preconditions is visible, skip the artifact rules and solve the question with the ordinary code-agent procedure. Do not spend an extra round trying to manufacture an intervention.}
\promptitem{0.6}{\textbullet}{Preserve the literal answer program: returned entity, population, filters, grouping unit, ordering, tie rule, selected field, units, aggregation, and rounding.}
\promptheading{Operation Boundary}
\promptitem{0.6}{\textbullet}{For each verified answer-relevant problem choose one outcome:}
\promptitem{1.7}{\textbullet}{\textbf{EDIT} only when independently usable visible evidence determines one exact replacement;}
\promptitem{1.7}{\textbullet}{\textbf{DROP} only when the record is demonstrably unusable for the question and no unique replacement is justified;}
\promptitem{1.7}{\textbullet}{\textbf{NO-OP} when the value is usable, irrelevant, unusual but valid, or supported only ambiguously.}
\promptitem{0.6}{\textbullet}{Candidate recurrence across rotations changes inspection priority, not authorization.}
\promptitem{0.6}{\textbullet}{Apply supported interventions before filtering, selection, ranking, adjacency, grouping, aggregation, or rounding, then rerun the unchanged answer program from the beginning.}
\promptheading{Missingness}
\promptitem{0.6}{\textbullet}{Activate only for a null, blank, whitespace-only string, or explicit missing marker in a field consumed by the answer program or required to validate it.}
\promptitem{0.6}{\textbullet}{Inspect returned values, filters, selectors, rank/order fields, ratio operands, and only their necessary companions. Do not bulk-fill unrelated columns.}
\promptitem{0.6}{\textbullet}{Test these visible recovery routes:}
\promptitem{1.7}{\textbullet}{an exact same-row total/parts, difference, product, ratio, average, or start/end/duration identity;}
\promptitem{1.7}{\textbullet}{an equivalent machine/display field or redundant representation;}
\promptitem{1.7}{\textbullet}{a complete-key mapping with exactly one value throughout comparable rows.}
\promptitem{0.6}{\textbullet}{Verify a proposed relation on every row where target and sources coexist. Reject it when a required source is missing, malformed, or contradictory.}
\promptitem{0.6}{\textbullet}{Use an alternate representation only to replace a missing answer-active value. Preserve every observed usable value in the primary question-specified field; never recompute the entire column from its alternate representation.}
\promptitem{0.6}{\textbullet}{EDIT only the missing field with its uniquely implied value.}
\promptitem{0.6}{\textbullet}{DROP only when the unrecoverable field is required to decide participation or contribution. Use NO-OP when it cannot affect the answer.}
\promptheading{Bad Values}
\promptitem{0.6}{\textbullet}{Activate only for a spreadsheet error, impossible token, explicit sentinel, or value that fails an answer-active field's necessary type or exact visible relation.}
\promptitem{0.6}{\textbullet}{Numeric sentinels remain numeric after \promptinline{pd.to\_numeric}; detect and mask an established sentinel explicitly before testing missingness or recovery.}
\promptitem{0.6}{\textbullet}{Establish a sentinel from its visible role and repeated conflict with a necessary relation, not merely from being negative, large, rare, or zero-padded.}
\promptitem{0.6}{\textbullet}{Preserve identifiers, codes, names, labels, titles, URLs, descriptions, categories, types, schools, and locations even when they contain digits or unusual text.}
\promptitem{0.6}{\textbullet}{Before DROP, test all applicable exact recovery routes:}
\promptitem{1.7}{\textbullet}{a redundant machine/display or category/code representation;}
\promptitem{1.7}{\textbullet}{a same-row total/parts, product/factors, rate/base, annual/hourly, or start/end/duration identity;}
\promptitem{1.7}{\textbullet}{a date-derived field or selector fixed by the question's grouping and ordering;}
\promptitem{1.7}{\textbullet}{a complete-key mapping with one value throughout comparable rows.}
\promptitem{0.6}{\textbullet}{Validate each relation on unaffected rows and verify every source. If one route yields exactly one value, EDIT before any query operation.}
\promptitem{0.6}{\textbullet}{DROP only when an answer-active record remains unusable and no independent anchor survives. Use NO-OP when evidence or repair direction is ambiguous.}
\promptheading{Inconsistent Formatting}
\promptitem{0.6}{\textbullet}{Activate only when representation blocks parsing or comparison of an answer-relevant value while its semantic value remains recoverable.}
\promptitem{0.6}{\textbullet}{Inspect the complete set of distinct answer-active token patterns before writing the parser; do not infer a delimiter or wrapper from one example.}
\promptitem{0.6}{\textbullet}{Match parsing to field role: numeric wrappers for measures, ordinal parsing for ranks, dates for time fields, and categories only when visible redundant fields establish the mapping.}
\promptitem{0.6}{\textbullet}{Parse the complete token with a role-specific pattern. Do not use the first numeric substring when another digit, sign, decimal mark, exponent, percent, or unit remains unexplained.}
\promptitem{0.6}{\textbullet}{Preserve every digit, sign, decimal position, exponent, percentage interpretation, and declared unit scale.}
\promptitem{0.6}{\textbullet}{Remove currency marks, grouping separators, units, percent notation, or text wrappers only when the complete token has one interpretation.}
\promptitem{0.6}{\textbullet}{Parse ordinal or wrapped rank values before filtering the requested ranks. Do not silently coerce malformed rank tokens to \promptinline{NaN} and drop them.}
\promptitem{0.6}{\textbullet}{Collapse one duplicated decimal mark only when that deletion yields a unique valid number matching the column's observed grammar. Never invent a decimal position.}
\promptitem{0.6}{\textbullet}{Edit only visibly formatted cells. Do not rewrite an already usable column merely to change its Python dtype.}
\promptitem{0.6}{\textbullet}{Standardize a category only when visible variants clearly denote the same entity and a canonical form is already present. Formatting alone never justifies DROP.}
\promptheading{Multi-Turn Execution Contract}
\promptitem{0.6}{\textbullet}{Keep canonical \promptinline{df} read-only. Use parsed Series, masks, or a local \promptinline{work\_df = df.copy()}.}
\promptitem{0.6}{\textbullet}{Prefer one compact inspection command that prints the exact active columns, malformed token patterns, and required relation diagnostics, followed by one complete answer command.}
\promptitem{0.6}{\textbullet}{Use full-table masks and original DataFrame indices; never copy indices from a sorted, filtered, or truncated preview.}
\promptitem{0.6}{\textbullet}{Before submission, verify that parsing found the expected records and that the computation produced the requested finite scalar, entity, or list.}
\promptitem{0.6}{\textbullet}{A failed assertion, empty parse, diagnostic count, or exception message is not an answer. Correct the parser or computation and execute it again.}
\promptitem{0.6}{\textbullet}{Once a valid result is observed, submit it immediately with the official \promptinline{done(answer)} command; do not spend later rounds re-deriving it.}

%% file: prompts/typeset/qa_base_system.tex
SETTING: You are an expert-level data scientist. Your job is to answer a data driven question   in rigorous manner given a data table.\par
In your analysis:\par
\promptitem{0.6}{\textbullet}{Carefully address}
\promptitem{0.6}{1)}{missing data: empty or null entries simulating incomplete information}
\promptitem{0.6}{2)}{bad values: clearly erroneous or placeholder entries (e.g., \promptinline{-1}, \promptinline{9999}, \promptinline{TEST}, \promptinline{\#REF!} etc.)}
\promptitem{0.6}{3)}{outliers: implausible extreme values that distort analysis (e.g., 220 breathing rate per minute)}
\promptitem{0.6}{4)}{inconsistent formatting: variations in representing the same value (e.g., \promptinline{22 lbs}, \promptinline{22 pounds}, \promptinline{weight = 22})}
\promptitem{0.6}{5)}{inconsistent logic: cross-field contradictions violating common-sense logic (e.g., end time before start time)}
\promptitem{0.6}{\textbullet}{Attempt to safely recover or correct flawed data when reasonable based on the existing data. If data is irrecoverable or suspect, discard the row.}
You will be working within a Python shell and can use the following commands to answer the question.\par
\promptheading{AVAILABLE COMMANDS:}
\promptcodefile{prompts/typeset/qa_base_system_code1.txt}
\promptcodefile{prompts/typeset/qa_base_system_code2.txt}
Please enclose all command kwargs values in \promptinline{\char34}\promptinline{\char34}\par
\tcbbreak
\promptheading{RESPONSE\_FORMAT:}
Each response must include:\par
\promptitem{0.6}{1.}{A DISCUSSION field — where you will methodically break down the reasoning process, illustrating how you arrive at conclusions and decide what to do next.}
\promptitem{0.6}{2.}{A command field — properly formatted YAML within triple backticks and following the structure from COMMANDS.}
Important rules:\par
\promptitem{0.6}{\textbullet}{Always include exactly one DISCUSSION and one command block.}
\promptitem{0.6}{\textbullet}{Ensure the command block is properly formatted YAML with proper indents and newlines (see the example below).}
For example, given a question asking for the average income. You might respond:\par
\promptheading{DISCUSSION}
Let's think step by step.   We need to first find the average income of the population.   We can do this by summing up the income column and dividing by the number of rows.\par
\promptcodefile{prompts/typeset/qa_base_system_code3.txt}

%% file: prompts/typeset/intervention_preturn.tex
The first assistant turn is reserved for query-conditioned intervention. Do not answer the downstream question and do not return \promptinline{done} in this turn. Return exactly one ordinary YAML \promptinline{python} command.\par
The Python namespace already contains the canonical object-typed \promptinline{df} and a\par
separate \promptinline{work\_df = df.copy()}. In one executable block:\par
\promptitem{0.6}{1.}{identify the literal answer program required by the question;}
\promptitem{0.6}{2.}{re-check only query-relevant ledger candidates and any directly necessary companion cells against the complete visible table;}
\promptitem{0.6}{3.}{decide \promptinline{EDIT}, \promptinline{DROP}, or \promptinline{NO-OP} under the operation boundaries;}
\promptitem{0.6}{4.}{apply every justified EDIT or DROP to \promptinline{work\_df}, leaving \promptinline{df} unchanged;}
\promptitem{0.6}{5.}{set \promptinline{intervention\_log} to a list of concise dictionaries containing the candidate coordinate, visible evidence, chosen action, and reason;}
\promptitem{0.6}{6.}{print a compact audit summary, but do not compute or print the final answer.}
An empty modification set is valid only after the applicable candidates have been checked; in that case retain \promptinline{work\_df} unchanged and record the relevant NO-OP decisions. The next user turn will contain the Python observation and open final QA. The same message history and Python namespace will persist.\par

%% file: prompts/typeset/qa_user.tex
Begin!\par
\promptfield{Data table (stored in a pandas dataframe named \promptinline{df}):}{[CANONICAL TABLE]}
All cells in the \promptinline{df} are \promptinline{object} data type, regardless of their appearance.\par
\promptfield{Question:}{[QUESTION]}
{\color{promptaccent}[RENDERED LEDGER]}\par
\promptheading{BEGIN RESERVED INTERVENTION TURN}
Perform the intervention preturn now. Do not answer the question in this turn.\par

%% file: prompts/typeset/qa_continuation.tex
Continue as the same agent in the same context. Answer the original question using the persistent \promptinline{work\_df} and the unchanged literal answer program. The intervention ledger is fallible: retain only operations justified by visible evidence. You may use ordinary Python inspection, then submit exactly one final answer with the official \promptinline{done(answer)} command. Do not export a repaired table.\par